\documentclass{article}

\PassOptionsToPackage{numbers, compress}{natbib}

\usepackage[preprint]{neurips_2025}

\usepackage[utf8]{inputenc} 
\usepackage[T1]{fontenc}    
\usepackage{hyperref}       
\usepackage{url}            
\usepackage{booktabs}       
\usepackage{amsfonts}       
\usepackage{nicefrac}       
\usepackage{microtype}      
\usepackage{xcolor}         
\usepackage{colortbl}

\usepackage{multirow} 
\usepackage{lscape} 
\usepackage{pdfpages} 
\usepackage{longtable} 
\usepackage[T1]{fontenc} 
\usepackage{graphicx} 
\usepackage{CJKutf8} 
\usepackage[most]{tcolorbox} 
\usepackage{amsmath}
\usepackage{amssymb}
\usepackage{mathtools}
\usepackage{amsthm}
\makeatletter

\newcommand{\Rmnum}[1]{\expandafter\@slowromancap\romannumeral #1@}
\makeatother
\usepackage{enumitem}

\title{Realistic Evaluation of TabPFN v2 in Open Environments}

\begin{document}

\author{%
  Zi-Jian Cheng$^{1,2}$,  Zi-Yi Jia$^{1,2}$,  Zhi Zhou$^{2,3}$,  Yu-Feng Li$^{2,3}$\thanks{Corresponding author.},  Lan-Zhe Guo$^{1,2}$\footnotemark[1] \\
$^1$School of Intelligence Science and Technology, Nanjing University, China\\
$^2$National Key Laboratory for Novel Software Technology, Nanjing University, China\\
$^3$School of Artificial Intelligence, Nanjing University, China\\
  \{chengzj,zhouz,liyf,guolz\}@lamda.nju.edu.cn,
jiazy@smail.nju.edu.cn \\
}

\maketitle
\vskip -0.3in
\begin{abstract}
Tabular data, owing to its ubiquitous presence in real-world domains, has garnered significant attention in machine learning research. While tree-based models have long dominated tabular machine learning tasks, the recently proposed deep learning model TabPFN v2 has emerged, demonstrating unparalleled performance and scalability potential. Although extensive research has been conducted on TabPFN v2 to further improve performance, the majority of this research remains confined to closed environments, neglecting the challenges that frequently arise in open environments. This raises the question: \textbf{Can TabPFN v2 maintain good performance in open environments?} To this end, we conduct the first comprehensive evaluation of TabPFN v2's adaptability in open environments. We construct a unified evaluation framework covering various real-world challenges and assess the robustness of TabPFN v2 under open environments scenarios using this framework. Empirical results demonstrate that \textbf{TabPFN v2 shows significant limitations in open environments but is suitable for small-scale, covariate-shifted, and class-balanced tasks}. Tree-based models remain the optimal choice for general tabular tasks in open environments. To facilitate future research on open environments challenges, we advocate for \textbf{open environments tabular benchmarks, multi-metric evaluation, and universal modules to strengthen model robustness}. We publicly release our evaluation framework at \href{https://anonymous.4open.science/r/tabpfn-ood-4E65}{the URL}.
\end{abstract}

\section{Introduction}
Tabular data~\cite{altman2017tabular} constitutes a highly structured data paradigm characterized by its organization of information through orthogonal dimensions of rows and columns~\cite{sahakyan2021explainable}. In tabular data, each row represents an instance, while each column encodes a specific feature or attribute. The pervasive applicability of tabular data has been demonstrated across diverse domains. Within financial services, it facilitates critical operations such as credit scoring~\cite{west2000neural} and quantitative portfolio management~\cite{zhu2021tat} through predictive analytics. In biomedical research, tabular datasets underpin clinical decision support systems~\cite{yildiz2024gradient} and pharmacological discovery pipelines~\cite{meijerink2020uncertainty}. To fully exploit the potential of tabular data for addressing real-world tasks, various tabular machine learning models have been developed. This evolutionary progression spans from tree-based methods (e.g., CatBoost~\cite{prokhorenkova2019catboostunbiasedboostingcategorical} and XGBoost~\cite{10.11452939672.2939785}) to deep learning models (e.g., ModernNCA~\cite{ye2024modernneighborhoodcomponentsanalysis} and TabPFN~\cite{Hollmann2022TabPFNAT, Hollmann2025AccuratePO}). These models have demonstrated exceptional performance across diverse tabular tasks.

Tree-based models consistently outperform deep learning models in tabular tasks \cite{grinsztajn2022tree, 13666122}. The emergence of a new deep learning model, TabPFN v2, has effectively disrupted the dominance of tree-based models in performance~\cite{Hollmann2025AccuratePO}. Grounded in the Transformer\cite{vaswani2017attention}, TabPFN v2 achieves state-of-the-art results through large-scale pre-training on synthetic datasets, allowing direct deployment on downstream tasks without the need for fine-tuning. Notably, TabPFN v2 introduces a novel contextual learning framework that processes both labelled training data and unlabeled test samples in unified input pipelines. This hybrid training approach facilitates joint optimization of feature representation and class prediction through self-supervised alignment mechanisms. Empirical validation across diverse datasets demonstrates an unprecedented performance level of TabPFN v2 in tabular tasks.

\begin{figure*}
\begin{center}
\centerline{\includegraphics[width=\linewidth]{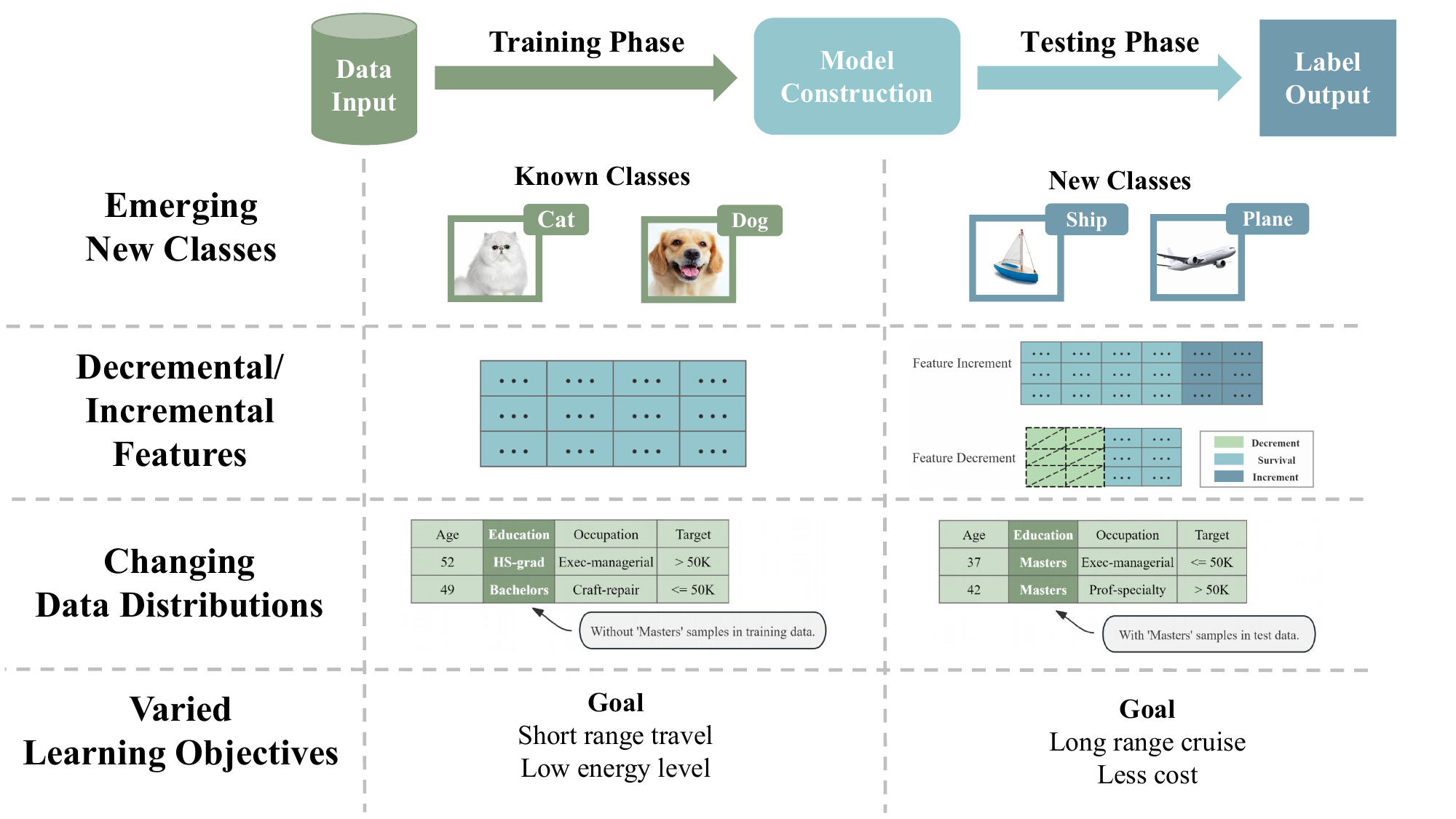}}
\vskip -0.1in
\caption{Open environments challenges in tabular data learning, including emerging new classes, decremental/incremental features, changing data distributions and varied learning objectives.}
\label{figure1}
\end{center}
\vskip -0.4in
\end{figure*}

Given the significant potential demonstrated by TabPFN v2 in handling tabular machine learning tasks, current research has focused on further enhancing its performance or adapting it to more real-world applications. They can be divided into two categories: performance evaluation and the handling of specific tasks. For performance evaluation,~\citet{liu2025tabpfnunleashedscalableeffective} had expanded the scope of evaluation experiments on TabPFN, assessing its performance on nearly 300 datasets and further validating the efficacy of TabPFN. To address the limitations of TabPFN v2 in handling high-dimensional, large-scale, and multi-class tabular machine learning tasks, a divide-and-conquer mechanism has been proposed~\cite{ye2025closerlooktabpfnv2}. Furthermore, researchers have proposed a series of optimization strategies to enhance TabPFN v2's adaptability in complex tasks such as context compression~\cite{koshil2024towards} and data generation~\cite{thomas2024retrieval}. ~\citet{koshil2024towards} suggests leveraging retrieval samples to construct a local context, thereby enhancing TabPFN v2's ability to perceive local information. While ~\citet{thomas2024retrieval} and~\citet{xu2025mixture} optimizes TabPFN v2's performance through data generation.

However, current research on TabPFN v2 is mostly carried out in closed environments where various learning factors, such as data distribution and feature space, remain consistent~\cite{parmar2023open}. In the real world, tabular tasks usually occur in open environments~\cite{zhou2022open} and face significant challenges when these learning factors change. For example, in traffic management systems, as the categories of traffic participants, event types, and facilities continue to increase, the complexity of management significantly rises \textbf{(Emerging New Classes)}. Meanwhile, equipment updates, failures, and changes in travel behaviour lead to feature drifts in data, affecting the system's accurate perception of traffic states \textbf{(Decremental/Incremental Features)}. Moreover, the distribution of traffic flow frequently changes due to factors such as urban planning, large-scale events, and holidays, further increasing the dynamism of management \textbf{(Changing Data Distributions)}. In addition, management goals have also shifted from single-efficiency optimization to multi-objective optimization, including reducing carbon emissions and enhancing system resilience, while paying more attention to long-term sustainability and overall system optimization \textbf{(Varied Learning Objectives)}. Figure~\ref{figure1} depicts four open environments challenges claimed in~\citet{zhou2022open}. Although existing research has gradually focused on improving TabPFN v2's adaptability in open environments, these studies mainly concentrate on distribution shift scenarios~\cite{hoo2025tabularfoundationmodeltabpfn, helli2024driftresilienttabpfnincontextlearning} and have not yet conducted a comprehensive evaluation of various challenges that TabPFN v2 may face in open environments. This limitation raises the natural question of \textbf{whether TabPFN v2 can maintain good performance in open environments}, and highlights the need for a more holistic assessment of TabPFN v2 in diverse and dynamic real-world scenarios.

To this end, we conduct a comprehensive evaluation of the performance of TabPFN v2 in open environments for the first time. Existing benchmarks for tabular data in open environments primarily evaluate models in isolated scenarios, limiting their methodological applicability to broader real-world tasks. To address this, we introduce a unified evaluation framework that systematically benchmarks diverse tabular models across various challenges in open environments, enabling standardized assessment of robustness and adaptability. 

From the experiments, we observe that \textbf{TabPFN v2 exhibits overall limitations across various challenges in open environments}. Although in emerging new classes, TabPFN v2 has the potential to detect new classes, when handling decremental/incremental features, it not only shows heightened vulnerability to feature decrement but also can not address newly added features during testing. Under changing data distributions, the performance of TabPFN v2 degrades substantially due to limited robustness against concept drift. For varied learning objectives, TabPFN v2 displays statistically significant bias toward majority classes while failing to maintain competitive performance across different task formulations. Moreover, the robustness of TabPFN v2 is fundamentally data-dependent, rendering its generalization capability highly sensitive to the scale of the dataset.

Although results demonstrate that tree-based models remain the optimal approach for general tabular tasks in open environments, the above observations suggest settings where TabPFN v2 is most likely the right choice in open environments for practitioners: 1) when the available dataset is small; 2) when the distribution shift is characterized as covariate shift; 3) when the label distribution is approximately balanced across classes.

Separately, state-of-the-art methods, despite their strong performance in closed environments, may fail to generalize effectively in open environments. This performance gap underscores a crucial research imperative regarding the enhancements required to advance open environments research. To address this challenge, we propose the following recommendations:

\begin{itemize}[leftmargin=*, labelsep=0.5em, nosep]
    \item Develop benchmarks targeting unexplored open environments tabular challenges.
    \item Evaluate models on various open environments metrics.
    \item Take model robustness as a critical metric when comparing model quality.
    \item Design universal modules to enhance the robustness of diverse existing models.
\end{itemize}

\section{Related Work}
\subsection{Open Environments Challenges}
Most tabular machine learning models are typically trained and tested in closed environments where critical learning factors remain stable. However, various real-world tasks operate in open environments where dynamic changes occur in key factors, posing challenges to model generalization~\cite{parmar2023open}.~\citet{zhou2022open} categorizes four core challenges in open environments: Emerging New Classes, Decremental/Incremental Features, Changing Data Distributions, and Varied Learning Objectives.

These challenges are pivotal in open environments machine learning. Emerging new classes, involving unseen classes during testing, have been addressed in natural language processing~\cite{NEURIPS2024_f6f4b34d} and computer vision~\cite{dhamija2020overlooked,du2022vos}. Decremental/Incremental features, caused by changes in feature sets, lead to mismatched training-testing spaces. TabFSBench~\cite{cheng2025tabfsbench} evaluates model performance under such variations, and~\citet{3294771} enhances performance by restoring ephemeral features. Changing data distributions, where test data violate the i.i.d.\ assumption, have led to benchmark datasets such as Tableshift~\cite{gardner2024benchmarking}, and methods such as domain adaptation~\cite{zhou2024fully} and domain generalization~\cite{zhou2021domaingeneralizationmixstyle}. Varied learning objectives, which prioritize adaptive optimization beyond accuracy, include multi-objective learning~\cite{zhou2019evolutionary,zuluaga2013active} and self-evolving training~\cite{liu2024diving}. However, research on these challenges remains fragmented, lacking a unified framework to evaluate models on all four challenges.

\subsection{Tabular Data in Machine Learning}
Tabular data, with structured and heterogeneous features, is used in healthcare, finance, and recommendation systems~\cite{Borisov_Leemann_Sessler_Haug_Pawelczyk_Kasneci_2022,kadra2021well,shwartz2022tabular}. Unlike images and texts, it has high dimensionality, heterogeneity, and complex dependencies, posing challenges for machine learning models~\cite{fang2024large}. Current approaches are mainly tree-based models (e.g., XGBoost~\cite{chizat2020faster}, LightGBM~\cite{badirli2020gradientboostingneuralnetworks}, CatBoost~\cite{prokhorenkova2019catboostunbiasedboostingcategorical}) and deep learning models. Tree-based models handle irregular patterns and uninformative features well~\cite{grinsztajn2022tree}, while deep learning models like DCN V2~\cite{wang2021dcn}, FT-Transformer~\cite{gorishniy2021revisiting}, and NODE~\cite{popov2019neuralobliviousdecisionensembles} aim to capture complex feature interactions for better performance~\cite{gorishniy2021revisiting,popov2019neuralobliviousdecisionensembles}.

In the realm of tabular machine learning tasks, tree-based models have traditionally held a dominant position over deep learning models~\cite{grinsztajn2022tree, 13666122}. The emergence of the novel deep learning model TabPFN v2~\cite{Hollmann2025AccuratePO} has surpassed tree-based models. TabPFN v2 has demonstrated superior performance compared to tree-based models across multiple benchmarks. However, the majority of these benchmarks are confined to closed environments. Consequently, the comparative performance of TabPFN v2 and tree-based models in open environments remains underexplored.

\subsection{Research on TabPFN}

TabPFN~\cite{Hollmann2022TabPFNAT}, short for Tabular Prior-Fitted Network, is a model pre-trained on large-scale synthetic datasets, enabling efficient zero-shot learning. It can efficiently perform classification and regression tasks without the need for hyperparameter tuning. Compared to existing models, TabPFN shows significant advantages on small- to medium-scale datasets with low computational cost, making it an efficient solution for tabular tasks. Recent research, however, reveals limitations in TabPFN's performance on high-dimensional, large-scale, or multi-class tasks~\cite{ye2025closerlooktabpfnv2, liu2025tabpfnunleashedscalableeffective}.

Various optimization strategies have been proposed to enhance TabPFN’s adaptability to current limitations and more complex scenarios, including local context construction via retrieval-based methods~\cite{koshil2024towards}, model fine-tuning~\cite{thomas2024retrieval,xu2025mixture}, and pretraining dataset expansion~\cite{breejen2024fine}. Moreover, TabPFN’s strong performance has prompted its application to challenges such as distribution shift adaptation~\cite{helli2024driftresilienttabpfnincontextlearning}, time series forecasting~\cite{hoo2025tabular}, and various domains including healthcare~\cite{noda2024machine,tran2024predicting}, ecology~\cite{heinzel2025advancing}, and cybersecurity~\cite{ruiz2024tabpfn}. However, these studies primarily focus on closed environments or target only a single challenge in open settings, lacking a comprehensive evaluation of TabPFN under diverse open environments scenarios.

\section{TabPFN and TabPFN v2}
This section explains how TabPFN and its newer version, TabPFN v2, work. Since there are already various research about these models, this section will give a short summary. More details are given in Appendix~\ref{tabpfn}, which brings together the important parts from~\citet{Hollmann2025AccuratePO,ye2025closerlooktabpfnv2}.

\subsection{TabPFN}  
Developed by~\citet{Hollmann2022TabPFNAT}, TabPFN reimagines classification through an innovative adaptation of a Transformer-based architecture. At its core, the method reformulates the classification task as a sequence processing problem with the following key components.

TabPFN standardizes each data point $(x_i, y_i)$ to $(\tilde{x}_i, \tilde{y}_i)$ in the $k$-dimensional space via linear projections with zero-padding ensuring uniform dimensionality. A context matrix $\mathcal{A}$ is constructed by concatenating $N$ training samples and a test sample $x^*$, where $\mathcal{A} = \left[ \tilde{x}_i \oplus \tilde{y}_i \right]_{i=1}^N \parallel \left[ \tilde{x}^* \right]$ and $\oplus$ denotes vector concatenation. This formulation treats each data point as a token in a sequence, enabling flexible handling of variable dataset sizes. The context matrix is then processed through Transformer layers and an MLP head, which converts the test sample's output token into class probabilities.

\subsection{TabPFN v2}  
Building upon TabPFN, TabPFN v2~\cite{Hollmann2025AccuratePO} introduces architectural innovations that redefine feature processing in tabular data analysis. The proposed method encompasses a feature space transformation where each raw feature is projected into a \( k \)-dimensional latent space and subjected to controlled perturbation, creating unique positional identifiers~\cite{ye2025closerlooktabpfnv2,gorishniy2021revisiting}. The computational framework processes a three-dimensional tensor structure using dual attention mechanisms: cross-sample attention for dataset-level patterns and intra-feature attention for feature relationships. Pre-trained weights derived from synthetic data generated by structural causal models which facilitate zero-shot transfer, thereby addressing the challenges of tabular data diversity.

Current research~\cite{Hollmann2025AccuratePO, ye2025closerlooktabpfnv2} has extensively evaluated TabPFN v2’s performance in closed environments, but largely overlooked its adaptability to open environments, leaving a critical gap. To fully realize its potential and practical value, we conduct comprehensive evaluations of TabPFN v2 under various open environments challenges.

\section{Open Environments Challenges}
\label{section4}
In this section, we draw upon the previous work presented in~\citet{zhou2022open} as a foundational framework to further symbolically formalize and represent the open environments challenges encountered in the tabular machine learning tasks. The detailed real-world descriptions are given in Appendix~\ref{challenge}.

\subsection{Emerging New Classes}
In closed environments machine learning tasks, it is commonly assumed that the class of any test sample must belong to the class set seen during training. However, this assumption does not always hold in open environments. We formally define this challenge by partitioning the class set \( L \) into \( L^{\text{train}} \) and \( L^{\text{test}} \), corresponding to the training and testing phases, respectively. In closed environments, the class set remains consistent between the training and testing phases, i.e., \( L^{\text{train}} = L^{\text{test}} \). In contrast, in open environments, test samples may belong to novel classes \( l \) that are not present during training, i.e., \( \exists ~l \in L^{\text{test}} \) such that \( l \notin L^{\text{train}} \). In such cases, the model must be capable of identifying and handling these new classes.

\subsection{Decremental/Incremental Features}
Decremental and incremental features represent open environments challenges characterized by partial removal or augmentation of the input feature set, known as feature shift. Let \(C\) denote the full feature set, partitioned into \(C^{\text{train}}\) and \(C^{\text{test}}\) for training and testing, respectively. In closed environments, \(C^{\text{train}} = C^{\text{test}}\), whereas in open environments, \(C^{\text{train}}\) remains fixed but \(C^{\text{test}}\) may differ. Specifically, when \(C^{\text{test}} \subsetneqq C^{\text{train}}\), imputation of shifted features in \(C^{\text{test}}\) is necessary to maintain input dimension consistency and enable accurate model prediction (\textbf{Decremental Features}). Conversely, when \(C^{\text{train}} \subsetneqq C^{\text{test}}\), the model typically truncates the newly added features in \(C^{\text{test}}\), retaining only those corresponding to \(C^{\text{train}}\), thus ensuring input dimension consistency between the training and testing phases (\textbf{Incremental Features}).

\subsection{Changing Data Distributions}
Closed environments machine learning research generally assumes that all data in both the training and testing phases are independent samples from the identical distribution. Unfortunately, this assertion does not always hold true in open environments. Changing data distributions has two scenarios. \textbf{Covariate Shift}~\cite{sugiyama2012machine} occurs when the input distribution \( p(x) \) changes between training and testing phases, while the conditional probability \( p(y|x) \) remains constant. \textbf{Concept Shift}~\cite{gama2014survey} involves changes in the conditional probability \( p(y|x) \) with a stable input distribution \( p(x) \).

\subsection{Varied Learning Objectives}
The performance of the machine learning model \( f \) can be measured by a learning objective \( M_f \), such as accuracy, F1 score, or ROC-AUC. Learning towards different objectives may lead to a model with different strengths. A model that is optimal on one measure is not necessarily optimal on others. Machine learning research in closed environments generally assumes that the \( M_f \) used to evaluate model performance is fixed and known in advance. However, this assumption may not always hold in open environments. When facing this challenge, the model should perform well across various learning objectives without requiring data to be recollected and a completely new model to be trained.

\subsection{Evaluation Framework for open environments Challenges}

Existing benchmarks for evaluating model performance in open environments typically focus on a single task, such as distribution shift~\cite{gardner2024benchmarking} or feature shift~\cite{cheng2025tabfsbench}. However, they lack a unified and comprehensive assessment across multiple open environments challenges. Hence, we propose a modular and extensible evaluation framework that assesses both model performance and robustness across diverse real-world scenarios. The framework formalizes four representative open environments challenges: Emerging New Classes, Decremental/Incremental Features, Changing Data Distributions, and Varied Learning Objectives. It builds testing protocols by leveraging existing benchmarks, including WhyShift~\cite{liu2024need} and TableShift~\cite{gardner2024benchmarking} for distribution shifts, and TabFSBench~\cite{cheng2025tabfsbench} for feature shifts. It supports comprehensive evaluation of tabular models and enables exporting datasets under different open environments scenarios with just a few lines of Python code. Details are in Appendix~\ref{website}.

\begin{table}[t]
\centering
\caption{Model performance on detecting new classes task. Higher values suggest stronger detection signals for emerging classes. The best is in \textbf{bold} and the second is \underline{underlined}.}
\label{tab:emerging_uncertainty_roc}
\resizebox{\textwidth}{15mm}{
\begin{tabular}{ccccccccc}
\toprule
\multirow{2}{*}{Model} & \multicolumn{2}{c}{EyeMovement} & \multicolumn{2}{c}{CMC}         & \multicolumn{2}{c}{Wine-Red}    & \multicolumn{2}{c}{Wine-White}  \\
\cmidrule(lr){2-3}\cmidrule(lr){4-5}\cmidrule(lr){6-7}\cmidrule(lr){8-9}
                       & \makebox[0.13\textwidth][c]{ROC-AUC}        & \makebox[0.13\textwidth][c]{AUPR}           & \makebox[0.13\textwidth][c]{ROC-AUC}        & \makebox[0.13\textwidth][c]{AUPR}           & \makebox[0.13\textwidth][c]{ROC-AUC}        & \makebox[0.13\textwidth][c]{AUPR}           & \makebox[0.13\textwidth][c]{ROC-AUC}        & \makebox[0.13\textwidth][c]{AUPR}            \\
                       \midrule
RandomForest           & 0.509          & \underline{0.505}    & 0.503          & 0.502          & 0.500          & 0.500          & 0.500          & 0.500          \\
XGBoost                & 0.503          & 0.502          & 0.502          & 0.501          & \underline{0.567}    & \underline{0.558}    & \underline{0.533}    & \underline{0.533}    \\
CatBoost               & 0.507          & 0.504          & 0.512          & 0.507          & 0.467          & 0.492          & 0.367          & 0.462          \\
MLP                    & 0.503          & 0.502          & \textbf{0.527} & \textbf{0.517} & \textbf{0.700} & \textbf{0.658} & 0.400          & 0.472          \\
RealMLP                & \underline{0.510}    & \underline{0.505}    & 0.514          & 0.508          & 0.500          & 0.500          & 0.500          & 0.500          \\
ModernNCA              & 0.504          & 0.502          & 0.510           & 0.506          & 0.400          & 0.481          & 0.500          & 0.500          \\
\rowcolor{gray!30} TabPFN v2              & \textbf{0.511} & \textbf{0.507} & \underline{0.522}    & \underline{0.513}    & 0.533          & 0.521          & \textbf{0.667} & \textbf{0.644}\\
\bottomrule
\end{tabular}}
\vskip -0.2in
\end{table}

\section{Comprehensive Evaluation of TabPFN v2}
Expanding on the impressive performance of TabPFN v2 in closed environments, we undertake a comprehensive evaluation in open environments to rigorously assess the robustness and adaptability of TabPFN v2 through our proposed evaluation framework. Specifically, we subject TabPFN v2 to evaluation across four distinct challenges in open environments, as detailed in Section~\ref{setting}. We choose RandomForest~\cite{breiman2001random}, XGBoost~\cite{10.11452939672.2939785} and CatBoost~\cite{prokhorenkova2019catboostunbiasedboostingcategorical} as tree-based baseline models. We also select MLP, RealMLP~\cite{holzmuller2025realmlp} and ModernNCA~\cite{ye2024modernneighborhoodcomponentsanalysis} as deep learning baseline models. Given the difference in datasets from different open environments challenges, we provide detailed descriptions of the datasets in each subsection. 

\subsection{Emerging New Classes}
Current tabular models (e.g., TabPFN v2) are fundamentally constrained by fixed input-output dimensions, limiting new class incorporation. To evaluate their adaptability, we design a novel class detection task from SMOOD~\cite{hendrycks2017baseline}. Based on multi-class datasets (Appendix~\ref{newclassdataset}), we implement a leave-one-class-out protocol: for $k$-class problems, we perform $k$ runs, each excluding one class during training and treating it as novel. We evaluate models on the metrics of Area Under the Precision-Recall curve (AUPR) and ROC-AUC. Results represent averages across all runs. The detailed computation procedure is described in Section~\ref{sec:newclass_results}.

\textbf{TabPFN v2 has the potential to detect new classes.}
As illustrated in Table~\ref{tab:emerging_uncertainty_roc}, TabPFN v2 consistently achieves better AUPR and ROC-AUC in the task of new class detection across all four datasets when compared to other models. This empirical evidence indicates that TabPFN v2 possesses a robust capability for identifying new classes. Results from other metrics of predicted probability are in Appendix~\ref{newclass}.

\subsection{Decremental/Incremental Features}
To conduct a comprehensive evaluation of decremental/incremental features, we adopt TabFSBench~\cite{cheng2025tabfsbench}, a benchmark specifically designed for this challenge. It includes twelve datasets covering eight classification tasks and four regression tasks across various domains, dataset sizes, and feature types. Descriptions and results are provided in Appendix~\ref{feature}.

\begin{table*}[t]
\caption{Average performance and performance gap across different tasks. We simulate feature shift by progressively altering the input feature set at five levels: 20\%, 40\%, 60\%, 80\%, and 100\%. For each level of shift, we compute the model's performance gap relative to the original (0\%) setting, separately for each task. For classification tasks, we report accuracy (higher is better), and for regression tasks, RMSE (lower is better). Each cell reports $x\,(\pm y)$, where $x$ denotes the performance under the given shift level, and $y$ represents the gap compared to the original performance. The best performance is shown in \textbf{bold}, and the smallest gap is \underline{underlined}.}
\vskip -0.2in
\label{table2}
\begin{center}
\begin{small}
\resizebox{\textwidth}{28mm}{
\begin{tabular}{cccccccccccc}
\toprule
Task                                   & Shift & RandomForest  & XGBoost       & CatBoost               & MLP                          & RealMLP       & ModernNCA              &\cellcolor{gray!30} TabPFN v2              \\
\midrule
\multirow{6}{*}{Binary Classification} & 0\%   & 0.838         & 0.842         & \textbf{0.869}         & 0.805                        & 0.813         & \textbf{0.869}         &\cellcolor{gray!30} 0.852                  \\
                                       & 20\%  & 0.764(-0.074) & 0.766(-0.076) & \textbf{0.834(-0.035)} & \underline{0.781(-0.024)}          & 0.744(-0.069) & 0.708(-0.161)          &\cellcolor{gray!30} 0.809(-0.043)          \\
                                       & 40\%  & 0.622(-0.216) & 0.624(-0.218) & \textbf{0.764(-0.105)} & \underline{0.743(-0.062)}          & 0.666(-0.147) & 0.598(-0.271)          &\cellcolor{gray!30} 0.725(-0.127)          \\
                                       & 60\%  & 0.583(-0.255) & 0.581(-0.261) & \textbf{0.714(-0.155)} & \underline{0.698(-0.107)}          & 0.672(-0.141) & 0.568(-0.301)          &\cellcolor{gray!30} 0.635(-0.217)          \\
                                       & 80\%  & 0.464(-0.374) & 0.514(-0.328) & \textbf{0.631(-0.238)} & \underline{0.620(-0.185)}          & 0.563(-0.250) & 0.540(-0.329)          &\cellcolor{gray!30} 0.556(-0.296)          \\
                                       & 100\% & 0.446(-0.392) & 0.467(-0.375) & \textbf{0.537(-0.332)} & \underline{0.534(-0.271)}          & 0.460(-0.353) & 0.512(-0.357)          &\cellcolor{gray!30} 0.483(-0.369)          \\
                                       \midrule
\multirow{6}{*}{Multi Classification}  & 0\%   & 0.800         & 0.802         & 0.837                  & 0.723                        & 0.745         & \textbf{0.906}         &\cellcolor{gray!30} 0.709                  \\
                                       & 20\%  & 0.735(-0.065) & 0.759(-0.043) & 0.794(-0.043)          & \underline{0.700(-0.023)}          & 0.640(-0.105) & \textbf{0.819(-0.087)} & \cellcolor{gray!30}0.651(-0.058)          \\
                                       & 40\%  & 0.637(-0.163) & 0.677(-0.125) & \textbf{0.714(-0.123)} & \underline{0.658(-0.065)}          & 0.665(-0.080) & 0.700(-0.206)          &\cellcolor{gray!30} 0.556(-0.153)          \\
                                       & 60\%  & 0.462(-0.338) & 0.574(-0.228) & \textbf{0.605(-0.232)} & \underline{0.600(-0.123)}          & 0.559(-0.186) & 0.562(-0.344)          &\cellcolor{gray!30} 0.432(-0.277)          \\
                                       & 80\%  & 0.354(-0.446) & 0.460(-0.342) & 0.463(-0.374)          & \underline{\textbf{0.520(-0.203)}} & 0.379(-0.366) & 0.444(-0.462)          &\cellcolor{gray!30} 0.288(-0.421)          \\
                                       & 100\% & 0.226(-0.574) & 0.306(-0.496) & 0.321(-0.516)          & \underline{\textbf{0.363(-0.360)}} & 0.195(-0.550) & 0.286(-0.620)          &\cellcolor{gray!30} 0.117(-0.592)          \\
                                       \midrule
\multirow{6}{*}{Regression}            & 0\%   & 0.925         & 0.922         & \textbf{0.902}         & 0.997                        & 0.926         & 0.940                  & \cellcolor{gray!30}0.928                  \\
                                       & 20\%  & 1.218(+0.293) & 1.155(+0.233) & 1.152(+0.250)          & \underline{1.025(+0.028)}          & 1.263(+0.337) & 1.103(+0.163)          & \cellcolor{gray!30}\textbf{0.974(+0.046)} \\
                                       & 40\%  & 1.537(+0.612) & 1.514(+0.592) & 1.544(+0.642)          & \underline{1.073(+0.076)}          & 1.567(+0.641) & 1.309(+0.369)          & \cellcolor{gray!30}\textbf{1.034(+0.104)} \\
                                       & 60\%  & 1.738(+0.813) & 1.762(+0.840) & 1.818(+0.916)          & \underline{1.125(+0.128)}          & 1.802(+0.876) & 1.499(+0.559)          & \cellcolor{gray!30}\textbf{1.184(+0.256)} \\
                                       & 80\%  & 2.086(+1.161) & 2.119(+1.197) & 2.247(+1.345)          & \underline{\textbf{1.181(+0.184)}} & 2.138(+1.212) & 1.735(+0.795)          & \cellcolor{gray!30}1.232(+0.304)          \\
                                       & 100\% & 2.346(+1.421) & 2.412(+1.490) & 2.571(+1.669)          & \underline{\textbf{1.247(+0.250)}} & 2.433(+1.507) & 1.940(+1.000)          & \cellcolor{gray!30}1.317(+0.389)         \\
\bottomrule
\end{tabular}}
\end{small}
\end{center}
\vskip -0.2in
\end{table*}

\textbf{TabPFN v2 exhibits heightened vulnerability to decremental features.} 
To assess TabPFN v2's adaptability to decremental features, we conduct random-shift experiments in TabFSBench and use the performance gap as a metric. The performance gap, explained in Appendix~\ref{feature}, measures the impact of feature shifts by comparing model performance metrics between original and shifted features. As shown in Tables~\ref{table2}, TabPFN v2's performance gap widens significantly with increasing feature shifts, indicating weaker adaptability and higher sensitivity to feature space changes. In contrast, MLP and CatBoost show greater robustness against decremental features, possibly due to their inherent anti-shift properties, which TabPFN v2 may lack.

\textbf{TabPFN v2 can not address new added features in the testing phase.}
When the dimensionality of input features dynamically increases, TabPFN v2 cannot process the additional features and can only truncate them to retain those present during training. This is because its internal parameters and feature representations are based on a fixed feature dimensionality during training. Consequently, TabPFN v2 cannot leverage the information from new features during testing. However, this limitation won't degrade TabPFN v2's performance.

\subsection{Changing Data Distributions}
\label{find3}
We evaluate TabPFN v2 under scenarios of changing data distributions, using metrics including Accuracy, Balanced Accuracy, F1-score, and ROC-AUC. Detailed results are given in Appendix~\ref{appendix:ood_results}. The evaluation is conducted on nine fully numerical datasets drawn from the WhyShift~\cite{liu2024need} and TableShift~\cite{gardner2024benchmarking} benchmarks, which contain three types of data distribution scenarios. To accommodate memory constraints and the current limitations of TabPFN v2 in handling very large datasets, we apply stratified subsampling (up to 50,000 instances) while preserving the original train/test splits. Detailed dataset statistics are provided in Appendix~\ref{dataset}. 

\textbf{TabPFN v2 reveals limited robustness when concepts shift.}
We present a comparative analysis of accuracy between XGBoost and TabPFN v2 under two distinct data distribution shifts: Concept Shift and Covariate Shift. XGBoost is the best model in the changing data distributions task. Figure~\ref{fig:acc} shows that both models exhibit enhanced accuracy when shifting from Concept to Covariate Shift. Meanwhile, XGBoost maintains consistent superiority over TabPFN v2 across both shift types. However, in Covariate Shift scenarios, TabPFN v2 demonstrates more improved performance compared to XGBoost, reducing the performance difference. Results on the other three metrics are given in Appendix~\ref{dd_result}. These results suggest that \textbf{TabPFN v2 demonstrates promising discriminative capacity on covariate-shift datasets instead of concept-shift datasets.}

\subsection{Varied Learning Objectives}
We conduct an exhaustive comparative analysis across four primary classification learning objectives: Accuracy, ROC-AUC, F1-score, and Balanced Accuracy. The analysis is performed on i.i.d. datasets used in changing data distribution.

\begin{figure*}
\begin{minipage}[t]{0.48\linewidth}
\centering
\includegraphics[width=0.85\linewidth]{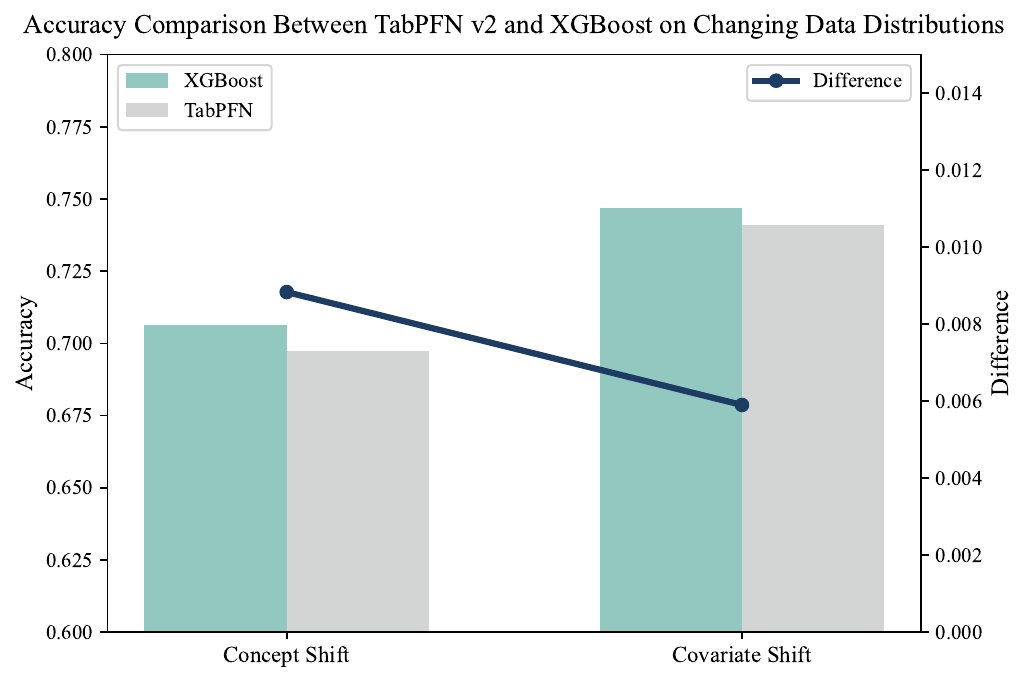}
\caption{Model performance comparison on accuracy between TabPFN v2 and XGBoost on changing data distributions task.}
\label{fig:acc}
\end{minipage}
\hspace{0.04\linewidth} 
\begin{minipage}[t]{0.48\linewidth}
\centering
\includegraphics[width=\linewidth]{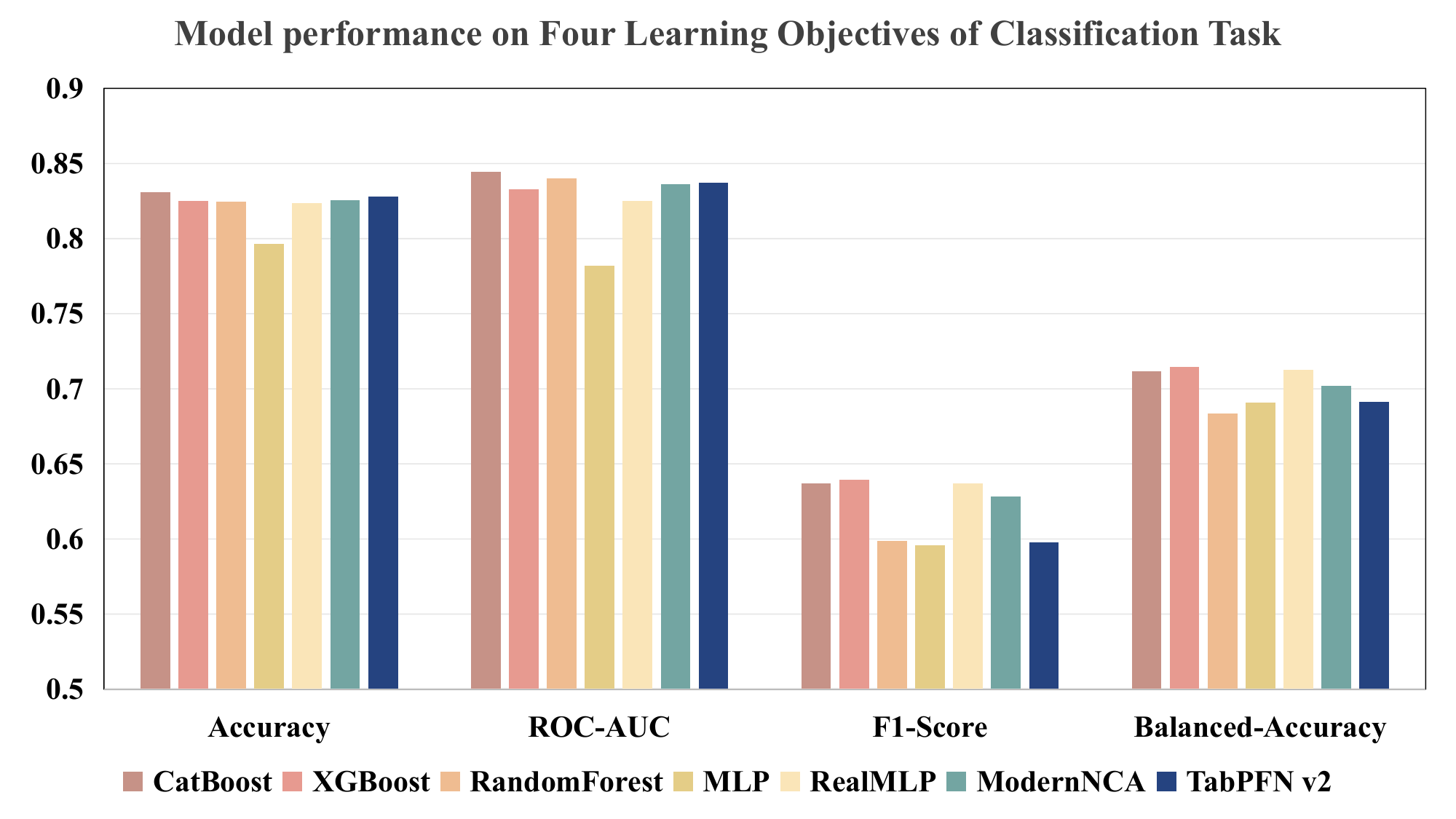}
\caption{Model performance on four learning objectives of classification task.}
\label{figure-t4}
\end{minipage}
\vspace{-5pt} 
\end{figure*}

\textbf{TabPFN v2 has statistically significant bias toward majority classes.}
Figure~\ref{figure-t4} reveals that the performance of TabPFN v2 degrades significantly on class-imbalance-sensitive metrics (F1-score and Balanced Accuracy), suggesting inherent limitations in handling minority classes. Specifically, Balanced Accuracy, a metric designed to address class imbalance by computing the arithmetic mean of per-class accuracies, shows that TabPFN v2 struggles to effectively adapt to varying sample sizes across different classes. Similarly, F1-score, as the harmonic mean of precision and recall, further confirms the model's suboptimal predictive capability for minority classes. Hence, \textbf{TabPFN v2 is suitable to handle datasets with balanced classes.}

\textbf{TabPFN v2 fails to maintain competitive performance across various learning objectives.}
As illustrated in Figure~\ref{figure-t4}, TabPFN v2 demonstrates competitive performance with respect to accuracy and ROC-AUC, achieving comparable results to other models in these specific evaluation criteria. However, a comparative analysis reveals statistically significant performance deficiencies in both F1 Score and Balanced Accuracy when contrasted with tree-based models and RealMLP. These empirical observations highlight an important limitation of TabPFN v2: while exhibiting superior performance for particular learning objectives, the model fails to maintain consistent efficacy across all evaluated performance metrics.

\subsection{Holistic Assessment}
We conduct a comprehensive assessment to evaluate the robustness of TabPFN v2 relative to compared models in open environments, employing a performance ranking analysis across four above challenges in open environments.

\textbf{TabPFN v2's robustness is inherently data-dependent.}
Through a thorough and comprehensive evaluation of the performance of TabPFN v2 across four distinct open environments challenges, our analysis has revealed that TabPFN v2 consistently demonstrates superior efficacy, particularly when applied to small-scale datasets. This observed performance characteristic is in precise alignment with the fundamental design objective of TabPFN v2, which is specifically optimized for small-scale datasets. The empirical results obtained substantiate the theoretical premise underlying TabPFN v2's development, thereby confirming \textbf{the particular suitability of TabPFN v2 for applications where the volume of training data is inherently limited}.

\textbf{Tree-based models remain the optimal approach for general tabular tasks in open environments.}
As shown in Table~\ref{tab:open_world_rankings}, tree-based models, particularly CatBoost and RandomForest, consistently outperform TabPFN v2. CatBoost achieves the best overall ranking, excelling in both changing data distributions and varied learning objectives, demonstrating stronger adaptability in open environments. In contrast, while TabPFN v2 remains competitive in closed environments, its performance declines relative to tree-based methods in open environments. These results suggest that tree-based models are better suited for open environments tasks requiring robustness.

\subsection{Recommendations}
During the experimental investigation, we observe that the majority of existing high-performance models predominantly demonstrate their superior performance in closed environments. However, these models tend to fall short in adapting to the open environments challenges that are more frequently encountered in real-world scenarios. To further enhance the performance of models in open environments and to provide guidance for the development of subsequent research, the following recommendations are proposed:

\textbf{Develop benchmarks targeting unexplored open environments tabular challenges.} Existing benchmarks are primarily designed around distribution shifts and feature shifts, lacking variations in open environments tabular challenges such as new classes and changes in learning objectives. Considering that the construction of benchmarks can effectively improve the model performance evaluation and methodological improvements on corresponding tasks. Therefore, it is urgent to develop benchmarks based on various open environments tabular challenges.

\textbf{Evaluate models on various open environments metrics.} Current research typically relies on OOD Accuracy, Performance Gap, or Balanced Accuracy to assess the robustness of a model. However, these metrics are mostly applicable to tasks involving distribution shifts or feature shifts and do not cover diverse open environments challenges. Therefore, additional general open environments metrics should be introduced in model evaluation, such as Open-World Tracking Accuracy~\cite{liu2022opening} and Mean Average Precision~\cite{sancaktar2022curious}.

\textbf{Take model robustness as a critical metric when comparing model quality.} Current research often judges the quality of a model solely based on its performance in closed environments, without considering the robustness of the model in open environments as an important evaluation criterion. While robustness is a crucial indicator for determining whether a model has practical value. Therefore, robustness should be regarded as a critical metric when comparing model quality, and the quality of a model should be comprehensively assessed based on both closed environments performance and open environments robustness.

\textbf{Design universal modules to enhance the robustness of diverse existing models.} From the aforementioned experiments, we learn that although some models perform well in certain open environments challenges, these models rely on their specific modules and lack universality. They cannot be transferred to other models to further improve robustness. Moreover, these models do not achieve excellent performance in all open environments challenges. Therefore, future research should focus on designing highly universal and transferable modules to enhance the overall performance of models in handling open environments tasks.

\begin{table*}[t]
\caption{Average rank across four open environments tasks. Ranks for emerging new classes are on ROC-AUC and AUPR. Ranks for decremental/incremental features are based on Accuracy and RMSE, while for changing data distributions and varied learning objectives are computed as the average rank across all metrics. The best rank is in \textbf{bold} and the second rank is \underline{underlined}.} 
\vskip -0.3in
\begin{center}
\resizebox{\textwidth}{12mm}{
\begin{tabular}{lccccccc}
\toprule
Task                             & RandomForest  & XGBoost    & CatBoost      & MLP  & RealMLP    & ModernNCA & \cellcolor{gray!30}TabPFN v2 \\
\midrule
Emerging New Classes             & 4.2 & 4.1 & 5.6 & 3.3 & \underline{3.2} & 5.2 & \cellcolor{gray!30}\textbf{1.8} \\
Decremental/Incremental Features & \textbf{1.78} & 4.89       & 3.89          & 4.00 & \underline{3.83} & 5.50      & \cellcolor{gray!30}4.06      \\
Changing Data Distributions      & 5.00          & \underline{2.25} & \textbf{2.00} & 6.00 & 4.25       & 4.25      & \cellcolor{gray!30}4.25      \\
Varied Learning Objectives       & 4.75          & \underline{2.75} & \textbf{2.00} & 6.75 & 4.00       & 3.75      & \cellcolor{gray!30}4.00      \\ 
\midrule
Average Rank                     & 3.93 & \underline{3.49} & \textbf{3.37} & 5.01       & 3.82      & 4.67& \cellcolor{gray!30} 3.53       \\
\bottomrule
\end{tabular}}
\end{center}
\label{tab:open_world_rankings}
\vskip -0.2in
\end{table*}

\section{Conclusion}
We present the first comprehensive evaluation of TabPFN v2 in open environments and construct an evaluation framework that simulates diverse challenges in open environments, revealing its limitations in feature decrements, and distribution shifts, while highlighting its strengths in detecting new classes, small-scale datasets and covariate shift scenarios. Although tree-based models remain superior for general tabular tasks, our analysis identifies specific conditions under which TabPFN v2 is pragmatically viable. The observations underscore a critical performance gap between closed and open environments, emphasizing the need for enhanced evaluation frameworks and robust model designs. To advance open environments research, we advocate for the development of specialized benchmarks, multi-faceted model assessments prioritizing robustness, and universal modules to improve existing methods’ adaptability. These directions aim to bridge the current methodological divide and foster more reliable tabular learning systems in real-world applications.

\textbf{Limitations.} Our experiments may not fully represent the diversity of open environments tasks due to constraints in dataset variety and task types, potentially impacting the simulation of complex real-world scenarios. The theoretical analysis depth may also limit insights into TabPFN v2's closed environments performance and open environments robustness.

\bibliography{main}
\bibliographystyle{plainnat}


\newpage
\appendix
\section{TabPFN and TabPFN v2}
\label{tabpfn}
\subsection{TabPFN}  
Developed by~\cite{Hollmann2022TabPFNAT}, TabPFN reimagines classification through an innovative adaptation of a Transformer-based architecture. At its core, the method reformulates the classification task as a sequence processing problem with the following key components:

\paragraph{Data Representation.}  
    Each data point $(x_i, y_i)$ undergoes $k$-dimensional standardization through linear projections:
    \[
    x_i \mapsto \tilde{x}_i \in \mathbb{R}^k, \quad y_i \mapsto \tilde{y}_i \in \mathbb{R}^k
    \]
    Where zero-padding ensures all vectors conform to the predefined dimensionality $k$.

\paragraph{Contextual Learning Framework.}  
    The model operates by constructing a dynamic context matrix $\mathcal{A}$ that jointly encodes with $N$ training samples and one test sample $x^*$:
    \[
    \mathcal{A} = \begin{bmatrix}
    \tilde{x}_1 \oplus \tilde{y}_1 \\
    \vdots \\
    \tilde{x}_N \oplus \tilde{y}_N \\
    \tilde{x}^*
    \end{bmatrix} \in \mathbb{R}^{(N+1) \times k}
    \]
    where $\oplus$ denotes vector concatenation. This formulation treats each transformed data point as a token in a sequence, enabling flexible handling of varying dataset sizes.

\paragraph{Architecture.}  
    The context matrix is processed through a stack of Transformer layers capable of processing variable-length token sequences and a specialized MLP head that converts the test instance's output token into class probabilities

The model's distinctive approach lies in its in-context learning paradigm, where the prediction for test sample $x^*$ emerges from the Transformer's processing of the entire augmented sequence containing both training and testing representations. This design eliminates the need for traditional iterative training while maintaining competitive accuracy on tabular tasks.

\subsection{TabPFN v2}  
Building upon TabPFN, TabPFN v2~\cite{Hollmann2025AccuratePO} introduces three key architectural innovations that redefine feature processing in tabular data analysis:

\paragraph{Feature Space Transformation.}  
    Each raw feature undergoes linear projection into a $k$-dimensional latent space, followed by controlled perturbation. This mechanism, characterized by~\cite{ye2025closerlooktabpfnv2} as a tokenization variant of~\cite{gorishniy2021revisiting}'s approach, creates unique positional identifiers for features.

\paragraph{Computational Framework.}  
    The computational framework operates on a three-dimensional tensor structure and processes through dual attention mechanisms of cross-sample attention for dataset-level patterns and intra-feature attention for feature relationships.

\paragraph{Knowledge Transfer.}  
    Pre-trained weights derived from synthetic data generated by structural causal model which facilitates zero-shot transfer, thereby addressing the challenges of tabular data diversity.

TabPFN v2 has three fundamental constraints: (1) quadratic complexity scaling, (2) dataset size limit $< 10^4$, and (3) maximum class count ($\leq 10$ for classification tasks). Hence,~\cite{ye2025closerlooktabpfnv2} introduces a divide-and-conquer mechanism to address these limitations. To address the performance degradation on high-dimensional datasets, a method combining feature subset sampling and ensemble learning is employed. For the inadequate performance on large-scale datasets, two improved schemes, data-to-embedding and decision tree, are proposed. To tackle the problem of inapplicability to multi-classification tasks, the Decimal Encoding and ECOC methods are utilized.

While current research~\cite{Hollmann2025AccuratePO, ye2025closerlooktabpfnv2} has thoroughly assessed TabPFN v2's performance, these evaluations primarily focus on its performance in closed environments. This leaves a critical gap in understanding how the model adapts to open environments. To fully realize TabPFN v2’s potential and explore its practical value, we conduct comprehensive evaluations. These evaluations focus on TabPFN v2’s performance under various open environments challenges.

\section{Tabular Challenges in Open Environments}
\label{challenge}
\subsection{Emerging New Classes}
In closed environments, it is commonly assumed that the label of any testing sample must come from the label set used during training. However, this assumption is not always valid in open environments. For instance, in a forest disease monitoring system that relies on a machine learning model trained with signals from sensors deployed in the forest, it is impractical to enumerate all possible classes in advance, as some forest diseases may be entirely novel, such as those caused by invasive insect pests that have never been encountered in the region before.

\subsection{Decremental/Incremental Features}
Decremental/Incremental Features are another open environments challenge, wherein the feature set previously utilized as inputs is either partially removed or expanded by new features, also known as feature shift. Given a forest disease monitoring system that relies on a machine learning model trained with signals from sensors deployed in the forest, certain existing sensors may cease to function, leading to a reduction of the feature set \textbf{(Decremental Features)}. Meanwhile, additional sensors may be deployed to monitor, resulting in an expansion of the feature set \textbf{(Incremental Features)}.

\subsection{Changing Data Distributions}
Machine learning research in closed environments generally assumes that all data in both the training and testing phases are independent samples from the identical distribution. Unfortunately, this assertion does not always hold true in open environments. In the forest disease monitoring system, the model may be built in summer based on sensor signals specific to that season, but it is expected to perform well across all seasons.

\subsection{Varied Learning Objectives}
The performance of the machine learning model \( f \) can be measured by a learning objective $M_f$, such as accuracy, F1 score, or ROC-AUC. Learning towards different objectives may lead to a model with different strengths. Being optimal on one measure does not mean that the model will also be optimal on other measures. Machine learning research in closed environments generally assumes that the $M_f$ used to measure model performance is invariant and known in advance. However, this assertion may not invariably be valid in open environments. In the forest disease monitoring system, the sensor dispatch task may prioritize different objectives over time. Initially, various sensors are dispatched to pursue high monitoring accuracy; later, after a relatively high accuracy has been achieved, different sensors may be used to ensure that the system operates with minimal energy consumption. When facing this challenge, the model should be able to perform well on various learning objectives without requiring the data to be recollected and a completely new model to be trained.

\section{Evaluation Framework}
\label{website}
To facilitate the use of our proposed evaluation framework, we provide a set of APIs. More details are available at \href{https://anonymous.4open.science/r/tabpfn-ood-4E65}{https://anonymous.4open.science/r/tabpfn-ood-4E65}. The API accepts four parameters: \texttt{dataset}, \texttt{model}, \texttt{task}, and \texttt{export\_dataset}. We will give further specifications in the supplementary material and the repository readme.md.

The \texttt{dataset} parameter specifies the full name of the dataset to be used. Our evaluation framework supports datasets from OpenML, Kaggle, and local directories.

The \texttt{model} parameter defines the model to be evaluated and can be selected from tree-based models and deep-learning models, which we evaluated in this paper. New models can be added by following the instructions in the "How to Add New Models" section.

The \texttt{task} parameter determines the type of feature-shift experiment to be conducted. The available options include emerging new classes (enc), decremental features (df), changing data distributions (cdd), and varied learning objectives (vlo).

The \texttt{export\_dataset} parameter controls whether the modified dataset—corresponding to a specific open environments challenge—is exported as a CSV file for further use.

An example command for running the evaluation framework is as follows ():

\vspace{10pt}
\begin{minipage}{\columnwidth}
\begin{tcolorbox}[colback=white, colframe=black, title=Example Command, fonttitle=\bfseries, rounded corners]
\texttt{python run.py ---dataset Adult ---model xgboost ---task enc ---export\_dataset True}
\end{tcolorbox}
\end{minipage}

\section{General Experimental Settings}
\label{setting}
\subsection{Traning settings}
Deep learning models are trained on an NVIDIA 4090 GPU. Tree-based models are trained on an AMD Ryzen 5 7500F 6-Core Processor. All experimental results are reported as the average of three different random seeds to ensure statistical reliability.

\subsection{Models}
In this subsection, we provide detailed descriptions of all the models used in our paper.

\paragraph{XGBoost} XGBoost\citep{chizat2020faster} is an efficient and flexible machine learning model that incrementally builds multiple decision trees by optimizing the loss function, with each tree correcting the errors of the previous one to continuously improve the model's predictive performance. XGBoost also incorporates the gradient boosting algorithm, iteratively training decision tree-based models with the goal of minimizing residuals and enhancing predictive accuracy.
\paragraph{CatBoost}
CatBoost~\cite{prokhorenkova2019catboostunbiasedboostingcategorical} is a powerful boosting-based model designed for efficient handling of categorical features. It uses the "Ordered Boosting" technique, which calculates gradients sequentially to prevent target leakage and maintain the independence of each training instance. At the same time, CatBoost employs "Target-based Categorical Encoding," converting categorical variables into numerical representations based on target statistics, thereby reducing the need for extensive preprocessing and improving model performance.
\paragraph{RandomForest} RandomForest~\cite{breiman2001random} is a classical ensemble learning method based on bagging and decision trees. It constructs a multitude of decision trees during training and outputs the mode or mean prediction of individual trees. Its robustness to overfitting, strong performance with minimal tuning, and ability to handle both classification and regression tasks make it a widely used baseline in tabular data benchmarks.
\paragraph{MLP} An MLP consists of multiple layers of neurons, with each layer fully connected to the next. An MLP contains at least three layers: an input layer, one or more hidden layers, and an output layer. It continuously adjusts the connection weights between neurons through training methods such as the backpropagation algorithm and gradient descent to minimize prediction errors.
\paragraph{ModernNCA} ModernNCA~\cite{ye2024modernneighborhoodcomponentsanalysis} is an enhanced Neighborhood Component Analysis (NCA) model that improves tabular data processing by adjusting learning objectives, integrating deep learning architectures, and using stochastic neighbor sampling for better efficiency and accuracy.
\paragraph{RealMLP} RealMLP~\cite{holzmuller2024better} is an enhanced multilayer perceptron designed for tabular data tasks, combining architectural improvements with meta-learned default hyperparameters. It achieves a strong balance between accuracy and training efficiency.

\subsection{Hyperparameter Tuning}
In this subsection, we provide hyperparameter grids of tree-based and deep learning models in Table~\ref{Hyperparameter_tree_model},~\ref{Hyperparameter_dl_model}. 

For tree-based models, we employ GridSearchCV from the scikit-learn library to conduct an exhaustive hyperparameter search. This approach systematically explores a predefined parameter grid through 5-fold cross-validation to ensure the reproducibility of results. The search process is optimized for computational efficiency by enabling parallel processing. 

Regarding deep learning models, we implement an adaptive hyperparameter optimization strategy based on the Optuna framework~\cite{akiba2019optuna}, following methodologies established in prior studies~\cite{liu2024talent}. The optimization protocol maintains a constant batch size of 1024 and performs 100 independent trials using training-validation splits to prevent potential data leakage from the test set.

\begin{table}[t]
\caption{Hyperparameter Grids of Tree-based Models. }
\begin{center}
\begin{tiny}
\label{Hyperparameter_tree_model}
\vskip 0.1in
\resizebox{0.8\textwidth}{!}{%
\begin{tabular}{ccc}
\toprule
\textbf{Model} & \textbf{Hyperparameter} & \textbf{Values} \\ 
\midrule
\multirow{6}{*}{\textbf{XGBoost}} 
& Learning Rate & $\{0.01, 0.1\}$ \\
& Max. Depth & $\{1, 5, 9\}$ \\
& N Estimators & $\{10000, 20000, 30000\}$ \\
& Subsample & $\{0.5, 0.8, 1.0\}$ \\
& Colsample Bytree & $\{0.5, 0.8, 1.0\}$ \\
& Min Child Weight & $\{1, 3, 5\}$ \\
\midrule
\multirow{3}{*}{\textbf{CatBoost}} 
& Learning Rate & $\{0.01, 0.05, 0.1\}$ \\
& Depth & $\{4, 6, 8\}$ \\
& Iterations & $\{500, 1000, 2000\}$ \\
\midrule
\multirow{2}{*}{\textbf{RandomForest}} 
& Min Samples Split & $[2,10]$ \\
& Min Samples Leaf & $[1, 10]$ \\
\bottomrule
\end{tabular}}
\end{tiny}
\end{center}
\vskip -0.1in
\end{table}

\begin{table}[!htbp]
\caption{Hyperparameter Grids of Deep Learning Models.}
\begin{center}
\begin{small}
\label{Hyperparameter_dl_model}
\vskip 0.1in
\resizebox{\textwidth}{!}{%
\begin{tabular}{ccc}
\toprule
\textbf{Model} & \textbf{Hyperparameter} & \textbf{Values} \\
\midrule
\multirow{4}{*}{\textbf{MLP}} 
& D\_layers & $\{1, 8, 64, 512\}$ \\
& Dropout & Uniform $\{0.0, 0.5\}$ \\
& Learning Rate & Loguniform$\{e^{-5}, 0.01\}$ \\
& Weight Decay & Loguniform$\{e^{-6}, 0.001\}$ \\
\midrule
\multirow{9}{*}{\textbf{ModernNCA}} 
& Dropout & Uniform $\{0.0, 0.5\}$ \\
& D\_block & Int$\{64, 1024\}$ \\
& N\_blocks & Int$\{0, 2\}$ \\
& N\_frequencies & Int$\{16, 96\}$ \\
& Frequency Scale & Loguniform$\{0.005, 10\}$ \\
& D\_embedding & Int$\{16, 64\}$ \\
& Sample Rate  & Uniform$\{0.05, 0.6\}$ \\
& Learning Rate & Loguniform$\{e^{-5}, 0.1\}$ \\
& Weight Decay & Loguniform$\{e^{-6}, 0.001\}$ \\
\midrule
\multirow{9}{*}{\textbf{RealMLP}} 
& Num Emb Type & $\{\text{none},\ \text{pbld},\ \text{pl},\ \text{plr}\}$ \\
& Add Front Scale & $\{\text{True},\ \text{False}\}$ \\
& Learning Rate (lr) & $\log U(0.02,\ 0.3)$ \\
& Dropout (p\_drop) & $\{0.00,\ 0.15,\ 0.30\}$ \\
& Activation (act) & $\{\text{selu},\ \text{relu},\ \text{mish}\}$ \\
& Hidden Sizes & $\{[256,256,256],\ [64,64,64,64,64],\ [512]\}$ \\
& Weight Decay (wd) & $\{0.0,\ 0.02\}$ \\
& PLR Sigma & $\log U(0.05,\ 0.5)$ \\
& Label Smoothing Epsilon (ls\_eps) & $\{0.0,\ 0.1\}$ \\
\bottomrule
\end{tabular}}
\end{small}
\end{center}
\vskip -0.1in
\end{table}

\section{Emerging New Classes}
\label{newclass}
\subsection{Dataset}
\label{newclassdataset}
\paragraph{Eye Movements}
This dataset is designed to predict the relevance of sentences in relation to a given question based on eye movement data. The target is to classify sentences as irrelevant, relevant, or correct, using 27 features, including landing position, first fixation duration, next fixation duration, time spent on the predicted region, and other relevant eye movement metrics. This dataset is available at \href{https://www.kaggle.com/datasets/vinnyr12/eye-movements}{https://www.kaggle.com/datasets/vinnyr12/eye-movements}.

\paragraph{Contraceptive Method Choice (CMC)}
This dataset contains 1,473 instances with 10 demographic and socio-economic attributes, originally derived from the 1987 National Indonesia Contraceptive Prevalence Survey. Each instance represents a married woman who was not pregnant (or unsure) at the time of the interview. The target is to predict the contraceptive method currently used by the individual, categorized into three classes: no-use, long-term methods, and short-term methods. This dataset was prepared by Tjen-Sien Lim and is available at \href{https://archive.ics.uci.edu/ml/datasets/Contraceptive+Method+Choice}{https://archive.ics.uci.edu/ml/datasets/Contraceptive+Method+Choice}.

\paragraph{Wine Quality (Red and White)}
This dataset includes two subsets related to red and white \textit{Vinho Verde} wine samples from the north of Portugal. Each sample is described by 11 physicochemical attributes (e.g., acidity, sugar, pH) and a quality score ranging from 1 to 4. The task is to predict the sensory quality of the wine based on its physicochemical properties. The dataset was first introduced by Cortez et al.~\cite{CorCer09} and is available at \href{https://archive.ics.uci.edu/dataset/186/wine+quality}{https://archive.ics.uci.edu/dataset/186/wine+quality}.

\subsection{Results}
\label{sec:newclass_results}
To evaluate a model's ability to detect novel classes, we adopt a leave-one-class-out protocol. For each class label, we exclude its samples from training and treat them as novel (label 1) during testing. An equal number of samples from the remaining classes are randomly sampled as known (label 0). After training the model on the reduced dataset, we compute confidence scores for test samples using the maximum predicted probability. Samples with low confidence (within a fixed interval $[\theta_{\min}, \theta_{\max}]$) are predicted as novel. We assess performance using ROC-AUC and AUPR, measuring the model’s ability to separate known and novel instances based on confidence.

To further assess whether models exhibit appropriate uncertainty when encountering novel classes, we hold out one or more classes during training and evaluate the predicted probabilities on test samples from these unseen classes. A prediction is deemed uncertain if its maximum confidence falls within a predefined low-confidence interval $[a, b]$. We report the proportion of novel samples falling into this interval as an indicator of the model's ability to recognize unfamiliar inputs. This metric complements ROC-AUC and AUPR by directly measuring how often the model expresses uncertainty when presented with out-of-distribution classes.

Table~\ref{tab:novel_class_detection}, Table~\ref{tab:emerging_uncertainty_045_055}, and Table~\ref{tab:emerging_uncertainty_049_051} show the results for uncertainty intervals $[0.4, 0.6]$, $[0.45, 0.55]$, and $[0.49, 0.51]$, respectively.

\begin{table}[t]
\centering
\caption{Proportion of predicted instances falling within the uncertainty interval $y_{\text{pred\_proba}} \in [0.4, 0.6]$ for the emerging new classes task. Higher values indicate stronger uncertainty responses to novel classes. The best is in bold and the second is \underline{underlined}.}
\label{tab:novel_class_detection}
\begin{tabular}{lcccc}
\toprule
Model        & EyeMovement    & CMC            & Wine-Red       & Wine-White     \\
\midrule
RandomForest & \textbf{0.716} & \underline{0.520}    & \textbf{0.764} & \textbf{0.562} \\
XGBoost      & 0.107          & 0.159          & 0.061          & 0.052          \\
CatBoost     & 0.150          & 0.183          & 0.127          & 0.070          \\
MLP          & 0.029          & 0.356          & 0.083          & 0.068          \\
RealMLP      & \underline{0.198}    & \textbf{0.523} & 0.000          & 0.000          \\
ModernNCA    & 0.101          & 0.323          & 0.182          & \underline{0.433}    \\
\rowcolor{gray!30} TabPFN v2      & 0.139          & 0.300          & \underline{0.246}    & 0.400     \\
\bottomrule
\end{tabular}
\end{table}

\begin{table}[t]
\centering
\caption{Proportion of detected new class data samples where the predicted probability $\in [0.45, 0.55]$, indicating uncertainty in classification. Higher values suggest stronger detection signals for emerging classes. The best is in \textbf{bold} and the second is \underline{underlined}.}
\label{tab:emerging_uncertainty_045_055}
\begin{tabular}{lcccc}
\toprule
Model & EyeMovement & CMC & Wine-Red & Wine-White \\
\midrule
RandomForest  & \textbf{0.340}  & \textbf{0.351}   & \textbf{0.244}  & \textbf{0.294} \\
XGBoost       & 0.054  & 0.075   & 0.027  & 0.023 \\
CatBoost      & 0.077  & 0.095   & 0.040  & 0.031 \\
MLP           & 0.014  & 0.203   & 0.037  & 0.052 \\
RealMLP       & \underline{0.091}  & \underline{0.272}   & 0.000  & 0.000 \\
ModernNCA     & 0.051  & 0.178   & 0.085  & 0.205 \\
\rowcolor{gray!30} TabPFN v2        & 0.072  & 0.156   & \underline{0.142}  & \underline{0.241} \\
\bottomrule
\end{tabular}
\end{table}

\begin{table}[t]
\centering
\caption{Proportion of predicted instances falling within the uncertainty interval $y_{\text{pred\_proba}} \in [0.49, 0.51]$ for the emerging new classes task. This narrow interval captures highly uncertain predictions. The best is in bold and the second is \underline{underlined}.}
\label{tab:emerging_uncertainty_049_051}
\begin{tabular}{lcccc}
\toprule
Model        & EyeMovement    & CMC            & Wine-Red       & Wine-White     \\
\midrule
RandomForest & \textbf{0.127} & \textbf{0.054} & \textbf{0.038} & 0.005          \\
XGBoost      & 0.012          & 0.016          & 0.001          & 0.001          \\
CatBoost     & 0.016          & 0.019          & 0.001          & 0.010          \\
MLP          & 0.003          & 0.038          & 0.003          & 0.004          \\
RealMLP      & \underline{0.019}    & \underline{0.053}    & 0.000          & 0.000          \\
ModernNCA    & 0.009          & 0.039          & 0.018          & \textbf{0.046} \\
\rowcolor{gray!30} TabPFN v2    & 0.015          & 0.036          & \underline{0.020}    & \underline{0.030}   \\
\bottomrule
\end{tabular}
\end{table}

\section{Decremental/Incremental Features}
\label{feature}
\subsection{Dataset}
We refer to TabFSBench~\cite{cheng2025tabfsbench} for details of the evaluated datasets.
\paragraph{Credit}
The original dataset contains 1,000 entries with 20 categorical/symbolic attributes prepared by Prof. Hofmann. In this dataset, each entry represents a person who takes a credit from a bank. Each person is classified as having good or bad credit risk according to the set of attributes. The target is to determine whether the customer's credit is good or bad. This dataset is available at \href{https://www.openml.org/search?type=data\&sort=runs\&id=31\&status=active}{https://www.openml.org/search?type=data\&sort=runs\&id=31\&status=active}.
\paragraph{Electricity}
The Electricity dataset, collected from the Australian New South Wales Electricity Market, contains 45,312 instances from May 1996 to December 1998. Each instance represents a 30-minute period and includes fields for the day, timestamp, electricity demand in New South Wales and Victoria, scheduled electricity transfer, and a class label. The target is to predict whether the price in New South Wales is up or down relative to a 24-hour moving average, based on market demand and supply fluctuations. This dataset is available on \href{https://www.kaggle.com/datasets/vstacknocopyright/electricity}{https://www.kaggle.com/datasets/vstacknocopyright/electricity}.
\paragraph{Heart}
Cardiovascular diseases (CVDs) are the leading cause of death globally, responsible for 17.9 million deaths annually.  Heart failure is a common event caused by CVDs, and this dataset contains 11 features that can be used to predict a possible heart disease. The target is to determine whether the patient's heart disease is present or absent. This dataset is available on \href{https://www.kaggle.com/datasets/fedesoriano/heart-failure-prediction
}{https://www.kaggle.com/datasets/fedesoriano/heart-failure-prediction
}.
\paragraph{Miniboone}  This dataset aims to construct a predictive model using various machine learning algorithms and document the end-to-end steps using a template. The MiniBooNE Particle Identification dataset is a binary classification task where we attempt to predict one of two possible outcomes. The target is to determine whether the neutrino is an electron or a muon. This dataset is available at \href{https://www.kaggle.com/datasets/alexanderliapatis/miniboone}{https://www.kaggle.com/datasets/alexanderliapatis/miniboone}.
\paragraph{Iris}
The Iris flower dataset, introduced by Ronald Fisher in 1936, contains 150 samples from three Iris species: Iris setosa, Iris virginica, and Iris versicolor. Each sample has four features: sepal length, sepal width, petal length, and petal width, measured in centimeters. The target is to classify the Iris species as setosa, versicolor, or virginica. This dataset is available on \href{https://www.kaggle.com/datasets/uciml/iris}{https://www.kaggle.com/datasets/uciml/iris}.
\paragraph{Jannis} This dataset is used in the tabular benchmark from~\cite{grinsztajn2022tree}. It belongs to the 'classification on numerical features' benchmark. The dataset is designed to test classification performance using numerical features, and it presents challenges such as varying data distributions, class imbalances, and potential missing values. It serves as a critical evaluation tool for machine learning models in real-world scenarios, including medical diagnosis, credit rating, and object recognition tasks. This dataset is available on \href{https://www.openml.org/search?type=data\&status=active\&id=45021}{https://www.openml.org/search?type=data\&status=active\&id=45021}.
\paragraph{Penguins}
Data were collected and made available by Dr. Kristen Gorman and the Palmer Station, Antarctica LTER, a member of the Long Term Ecological Research Network. The goal of the Palmer Penguins dataset is to offer a comprehensive resource for data exploration and visualization, serving as an alternative to the Iris dataset. The target is to classify the penguin species as Adelie, Chinstrap, or Gentoo. This dataset is available at \href{https://www.kaggle.com/datasets/youssefaboelwafa/clustering-penguins-species}{https://www.kaggle.com/datasets/youssefaboelwafa/clustering-penguins-species}.

\begin{table}[t]
\centering
\caption{Summary statistics of three selected datasets from TableShift after subsampling.}
\begin{center}
\begin{small}
\label{tab:tableshift_stats}
\resizebox{\textwidth}{!}{%
\begin{tabular}{lrrrrrrrrrr}
\toprule
\textbf{Dataset} & \textbf{Train Size} & \textbf{ID Test} & \textbf{OOD Test} & \textbf{Total} & \textbf{Num. Columns} & \textbf{Cat. Columns} & \textbf{\# Classes} & $\Delta_x(\text{Eqn.}~\ref{eqn:xshift})$ & $\Delta_{y|x}(\text{Eqn.}~\ref{eqn:dfd})$ & $\Delta_y(\text{Eqn.}~\ref{eqn:delta-y})$ \\
 \\
\midrule
college\_scorecard        & 43,908 & 5,488  & 602    & 49,998 & 118 & 0 & 2 & 43,566.39 & 2116.63  & 0.0337\\
brfss\_diabetes         & 37,284 & 4,660  & 8,054  & 49,998 & 142 & 0 & 2 & 12.28 & 0.10 & 0.0332\\
diabetes\_readmission     & 19,146 & 2,393  & 28,460 & 49,999 & 183 & 0 & 2 & 42.37 & 1.30 & 0.0060\\
\bottomrule
\end{tabular}%
}
\end{small}
\end{center}
\end{table}

\begin{table}[t]
\centering
\caption{Overview of selected WhyShift datasets and settings (after subsampling). The evaluation of Shift Pattern is described in \ref{sec:shift-contribution}.}
\label{tab:whyshift_stats}
\begin{center}
\begin{small}
\resizebox{\textwidth}{!}{%
\begin{tabular}{lrrrrrll}
\toprule
\textbf{Dataset} & \textbf{Train} & \textbf{ID Test} & \textbf{OOD Test} & \textbf{Total} & \textbf{\#Features} & \textbf{Setting} & \textbf{Shift Pattern} \\
\midrule
ACS Income (CA$\rightarrow$PR)           & 38,227  & 9,557  & 2,215   & 49,999 & 9  & California $\rightarrow$ Puerto Rico & $Y|X \gg X$ \\
ACS Mobility (MS$\rightarrow$HI)         & 4,254   & 1,064  & 2,733   & 8,051  & 21 & Mississippi $\rightarrow$ Hawaii    & $Y|X \gg X$ \\
ACS Pub.Cov (NE$\rightarrow$LA)          & 23,211  & 5,065  & 1,267   & 16,879 & 18 & Nebraska $\rightarrow$ Louisiana    & $Y|X \gg X$ \\
ACS Pub.Cov (2010$\rightarrow$2017)      & 20,501  & 5,126  & 24,372  & 49,999 & 18 & 2010 (NY) $\rightarrow$ 2017 (NY)    & $Y|X \ll X$ \\
ACS Income (Young 80\%)                  & 20,000  & 5,000  & 25,000  & 50,000 & 9  & Younger People (80\%)               & $Y|X \ll X$ \\
ACS Income (Young 90\%)                  & 20,000  & 5,000  & 25,000  & 50,000 & 9  & Younger People (90\%)               & $Y|X \ll X$ \\
\bottomrule
\end{tabular}%
}
\end{small}
\end{center}
\end{table}

\paragraph{Eye Movements}
This dataset is designed to predict the relevance of sentences in relation to a given question based on eye movement task. The target is to classify sentences as irrelevant, relevant, or correct, using 27 features, including landing position, first fixation duration, next fixation duration, time spent on the predicted region, and other relevant eye movement metrics. This dataset is available at \href{https://www.kaggle.com/datasets/vinnyr12/eye-movements}{https://www.kaggle.com/datasets/vinnyr12/eye-movements}.
\paragraph{Abalone}
The age of abalone is traditionally determined by cutting the shell, staining it, and counting the rings under a microscope, a process that is both tedious and time-consuming. This dataset uses easier-to-obtain physical measurements, such as length, diameter, and weight, to predict the abalone's age. The target is to predict the age, providing a more efficient approach.  This dataset is available on \href{https://www.kaggle.com/datasets/rodolfomendes/abalone-dataset}{https://www.kaggle.com/datasets/rodolfomendes/abalone-dataset}.
\paragraph{Bike}
The dataset records the rental of shared bikes in the Washington area from 2011-01-01 to 2012-12-31, containing 11 features such as season, holiday, working day, and weather conditions. The goal is to predict the total count of bikes rented each hour, with the target being to forecast the number of bicycles available for rent today based on historical rental patterns and external factors like temperature, humidity, and seasonal trends. This dataset is available on \href{https://www.kaggle.com/datasets/abdullapathan/bikesharingdemand}{https://www.kaggle.com/datasets/abdullapathan/bikesharingdemand}.
\paragraph{Concrete}
Concrete is the most important material in civil engineering, and its compressive strength is influenced by a highly nonlinear relationship with its ingredients and age. The dataset contains 9 attributes, including variables such as cement, water, and age. The target is to predict the concrete compressive strength (measured in MPa) using these input variables. This dataset is available on \href{https://www.kaggle.com/datasets/maajdl/yeh-concret-data}{https://www.kaggle.com/datasets/maajdl/yeh-concret-data}.
\paragraph{Laptop}
The original dataset was relatively compact, with many details embedded in each column. The columns mostly consisted of long strings of data, which were relatively human-readable and concise. However, for Machine Learning algorithms to work more efficiently, it is better to separate different details into individual columns. After doing so, 28 duplicate rows were exposed and removed. The cleaned dataset serves as the final result. The target is to predict the price of a laptop based on its specifications. This dataset is available on \href{https://www.kaggle.com/datasets/owm4096/laptop-prices}{https://www.kaggle.com/datasets/owm4096/laptop-prices}.

\subsection{Performance Gap}
We consider the percentage of model performance gap \textbf{$\Delta$} as model robustness in feature-shift scenarios by following TabFSBench~\cite{cheng2025tabfsbench}, 
\begin{equation}
\Delta  = \frac{(metric_i- metric_0 )}{metric_0}
\label{delta_equation}
\end{equation}
$metric_i$ denotes the model performance where $i$ features shift. In subsequent sections, we use $metric$ to refer to \textbf{performance}, and $\Delta$ to refer to \textbf{robustness}.

\subsection{Results}
To systematically assess TabPFN v2 in the presence of decremental features, we design random-shift experiments in TabFSBench. We use accuracy and ROC-AUC for classification tasks and RMSE for regression tasks. Table~\ref{table2} provides detailed model performance.

\section{Changing Data Distributions}
\label{appendix:ood_results}
\subsection{Dataset}
\label{dataset}
We evaluate our models on two established benchmarks for distribution shift in tabular data: \textbf{TableShift}\citep{gardner2024benchmarking} and \textbf{WhyShift}\citep{liu2024need}.

From the TableShift benchmark, we select three datasets—\textit{College Scorecard}, \textit{Hospital Readmission}, and \textit{Diabetes}. To ensure scalability and consistency across experiments, we apply stratified subsampling to limit each dataset to 50,000 total instances while preserving the original train/test split ratio. Detailed statistics are provided in Table~\ref{tab:tableshift_stats}.

From the WhyShift benchmark, we adopt six pre-defined settings provided by the original paper, covering a variety of real-world covariate and concept shift scenarios. For datasets containing more than 50,000 instances, we similarly apply stratified subsampling to retain a total of 50,000 samples. The configuration and statistics for all selected WhyShift settings are summarized in Table~\ref{tab:whyshift_stats}.

\subsection{Domain Shift Metrics}
Domain shift can be categorized into covariate shift, concept shift, and label shift. We adopt the metrics proposed in TableShift~\cite{gardner2024benchmarking} to quantify the degree of these three types of domain shift.

\paragraph{Measuring Covariate Shift with OTDD:} 
\begin{equation}
    \Delta_x = \text{OTDD}(\mathcal{D}^{\text{train}}, \mathcal{D}^{\text{test}})
    \label{eqn:xshift}
\end{equation}
Here, \(\mathcal{D}_{\text{train}}\) and \(\mathcal{D}_{\text{test}}\) denote the source and target domain datasets, respectively. OTDD represents the Optimal Transport Dataset Distance, computed under a Gaussian approximation~\cite{alvarez2020geometric}.

\paragraph{Measuring Concept Shift with Frechet Dataset Distance (FDD):}  
Inspired by the widely used Frechet Inception Distance (FID) in machine learning~\cite{heusel2017gans}, FDD utilizes intermediate representations of a classifier to quantify distributional discrepancies. It calculates the Frechet distance (also known as the Wasserstein-2 distance) between two distributions to assess the extent of concept shift.

The computation of this metric proceeds as follows: First, a classifier (we use MLPs) is trained on the source domain using the best hyperparameters obtained through hyperparameter search. Then, for each input \(x \in \mathcal{D}\), we compute the activation values at each layer of the model, obtaining the activation vector \(\hat{x} := f_\theta[i](x)\), where \(i\) denotes the \(i\)-th layer of the model. Finally, the Frechet Dataset Distance is calculated to measure the divergence between the two distributions.

\begin{equation}
\label{eqn:dfd}
    \textrm{DFD}(\mathcal{D}^{\text{train}}, \mathcal{D}^{\text{test}}) =  ||\mu_{\mathcal{D}^{\text{train}}} -\mu_{\mathcal{D}^{\text{test}}}||^2   + Tr(\Sigma_{\mathcal{D}^{\text{train}}} + \Sigma_{\mathcal{D}^{\text{test}}} - 2*\sqrt{\Sigma_{\mathcal{D}^{\text{train}}}*\Sigma_{\mathcal{D}^{\text{test}}}})
\end{equation}

\(\mu_{D}\) represents the set of feature vectors extracted from domain \(D\). Based on \(\mu_{D}\), we construct the covariance matrix \(\Sigma_{D}\). In the discussion below, we refer to the resulting measure as \(\Delta_{y|x}\). A lower FDD score indicates a smaller distance between any \(x_i\) in the training domain \(D_{\text{train}}\) and \(x_j\) in the test domain \(D_{\text{test}}\).

\begin{figure}[t]
  \centering
  \includegraphics[width=0.7\linewidth]{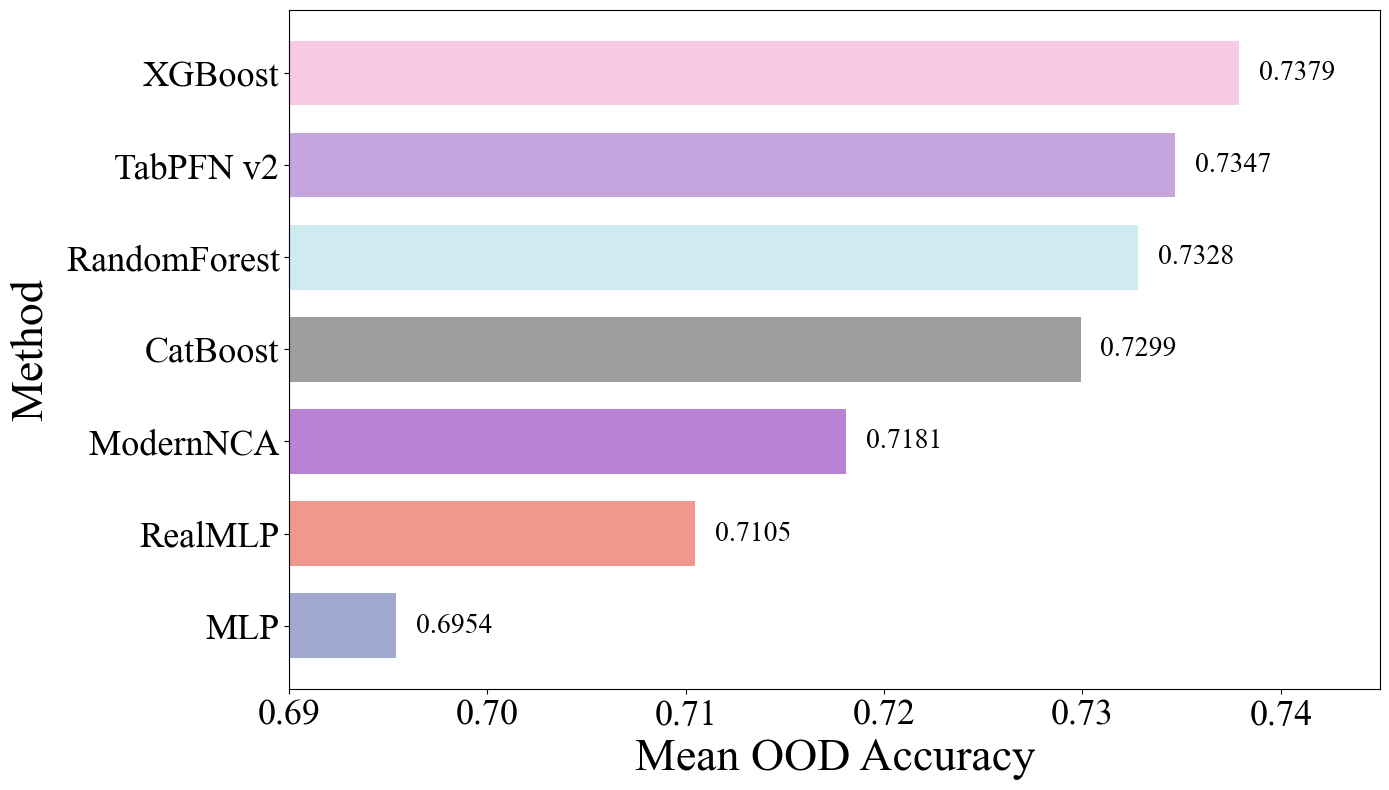}
  \caption{Mean Balance Accuracy Across 9 Datasets with Distribution Shifts}
  \label{image/acc.png}
\end{figure}

\begin{figure}[t]
  \centering
  \includegraphics[width=0.7\linewidth]{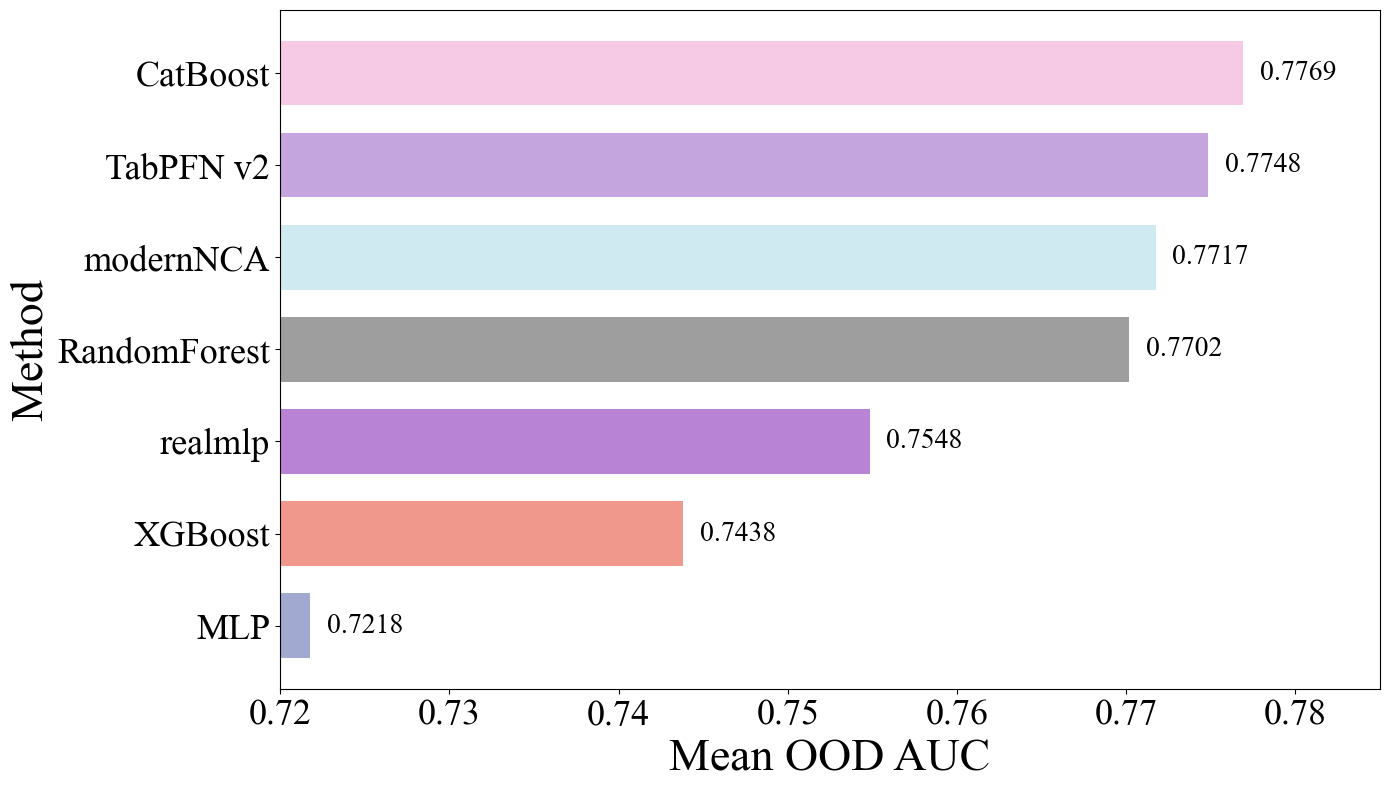}
  \caption{Mean AUC Across 9 Datasets with Distribution Shift}
  \label{fig:auc}
\end{figure}

\paragraph{Measuring label shift:}
TableShift~\cite{gardner2024benchmarking} proposes a simple formula to quantify the label shift between the source and target distributions:

\begin{equation}
    \Delta_y = ||\Bar{y}_{\mathcal{D}^{\text{train}}} - \Bar{y}_{\mathcal{D}^{\text{test}}}||^2
    \label{eqn:delta-y}
\end{equation}

In this equation, \(\Bar{y}_{\mathcal{D}} =  \frac{1}{|\mathcal{D}|} \sum_{i \in \mathcal{D}}y_i\) represents the average label value computed from samples within domain \(\mathcal{D}\). Given that all tasks in our study are binary classification, this formulation captures the squared \(L_2\) distance between the class prior probabilities of the source and target domains.

\paragraph{Quantifying the Contribution of \texorpdfstring{$X$}{X}-Shifts and \texorpdfstring{$Y|X$}{Y|X}-Shifts}
\label{sec:shift-contribution}

To gain a fine-grained understanding of the sources of model performance degradation under distribution shifts, we adopt the DIStribution Shift DEcomposition (DISDE) framework proposed by ~\cite{Cai_Namkoong_Yadlowsky_2023}. This framework enables the decomposition of the overall generalization gap into components attributed to covariate shifts ($X$-shifts) and conditional label shifts ($Y|X$-shifts).

Formally, for a model $f_P$ trained on a source distribution $P$ and evaluated on a target distribution $Q$, DISDE decomposes the generalization gap as:

\begin{equation}\label{disde}
\begin{aligned}
    \mathbb{E}_Q[\ell(f_P(X), Y)] - \mathbb{E}_P[\ell(f_P(X), Y)] 
    &= \mathbb{E}_{S_X}[R_P(X)] - \mathbb{E}_P[R_P(X)] \hfill &(\Rmnum{1})& \\
    &\quad + \mathbb{E}_{S_X}[R_Q(X) - R_P(X)] \hfill &(\Rmnum{2})& \\
    &\quad + \mathbb{E}_Q[R_Q(X)] - \mathbb{E}_{S_X}[R_Q(X)] \hfill &(\Rmnum{3})&
\end{aligned}
\end{equation}

where $R_{\mu}(x) = \mathbb{E}_{\mu}[\ell(f_P(X), Y) \mid X = x]$ denotes the conditional expected loss under distribution $\mu \in \{P, Q\}$, and $S_X$ is an auxiliary distribution over $X$ whose support is contained within both $P_X$ and $Q_X$.

Each term in \text{Eqn.}~\ref{disde} in this decomposition corresponds to a specific type of shift: \begin{itemize} 
\item \text{Eqn.}~\ref{disde}.\Rmnum{1}\&\Rmnum{3} reflect changes due to differences in the marginal distribution of covariates—i.e., the contribution of $X$-shifts. 
\item \text{Eqn.}~\ref{disde}.\Rmnum{2} captures the shift in the conditional distribution of labels given features, corresponding to $Y|X$-shifts. 
\end{itemize}

Building upon this decomposition, we utilize the open-source \texttt{WhyShift} package\footnote{\url{https://github.com/namkoong-lab/whyshift}}, which implements DISDE in a scalable and extensible manner. This allows us to rigorously quantify the relative impact of $X$-shifts and $Y|X$-shifts on performance degradation across datasets and domains, providing deeper insight into model robustness under open environments evaluation settings.

\subsection{Results}
\label{dd_result}
\begin{figure}[t]
  \centering
  \includegraphics[width=0.7\linewidth]{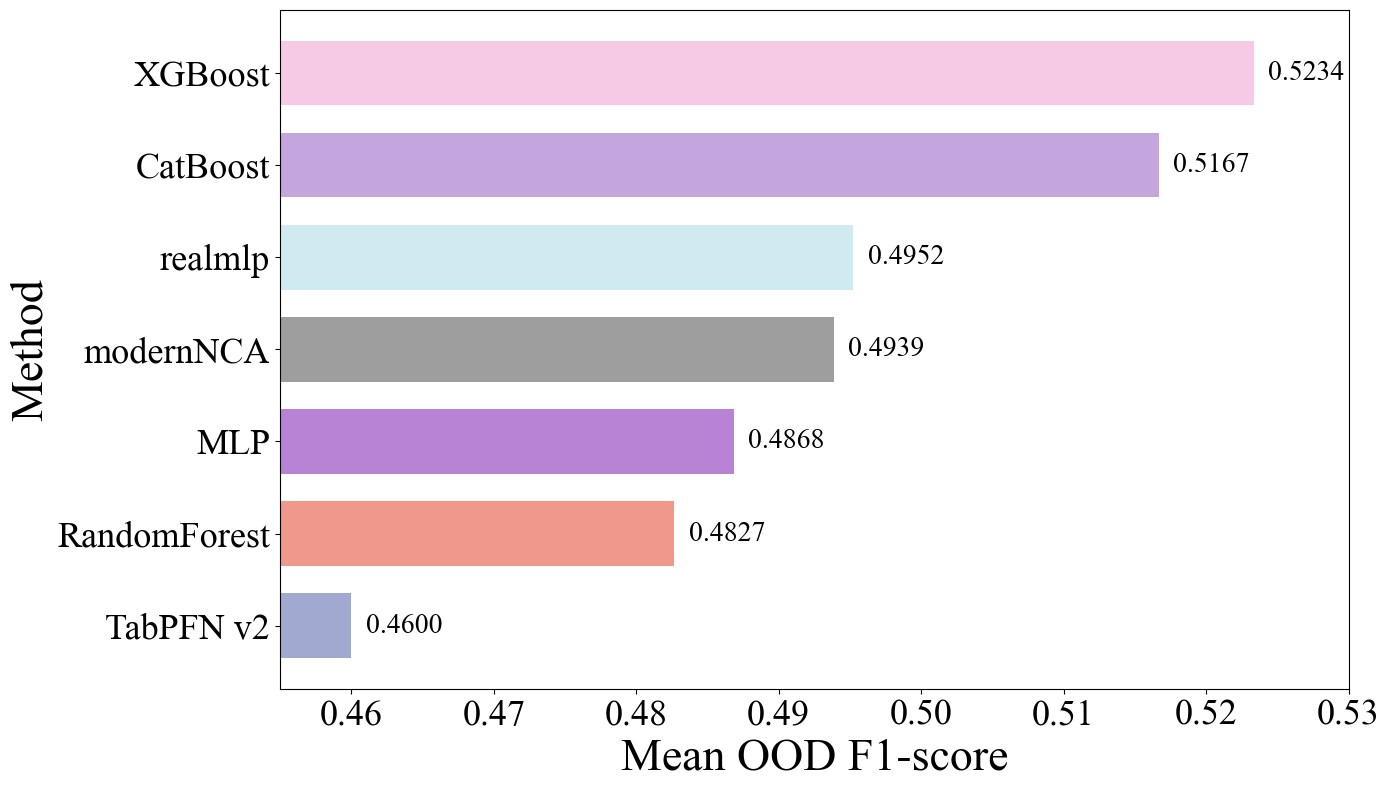}
  \caption{Mean F1-score Across 9 Datasets with Distribution Shift}
  \label{fig:f1}
\end{figure}

\begin{figure}[t]
  \centering
  \includegraphics[width=0.7\linewidth]{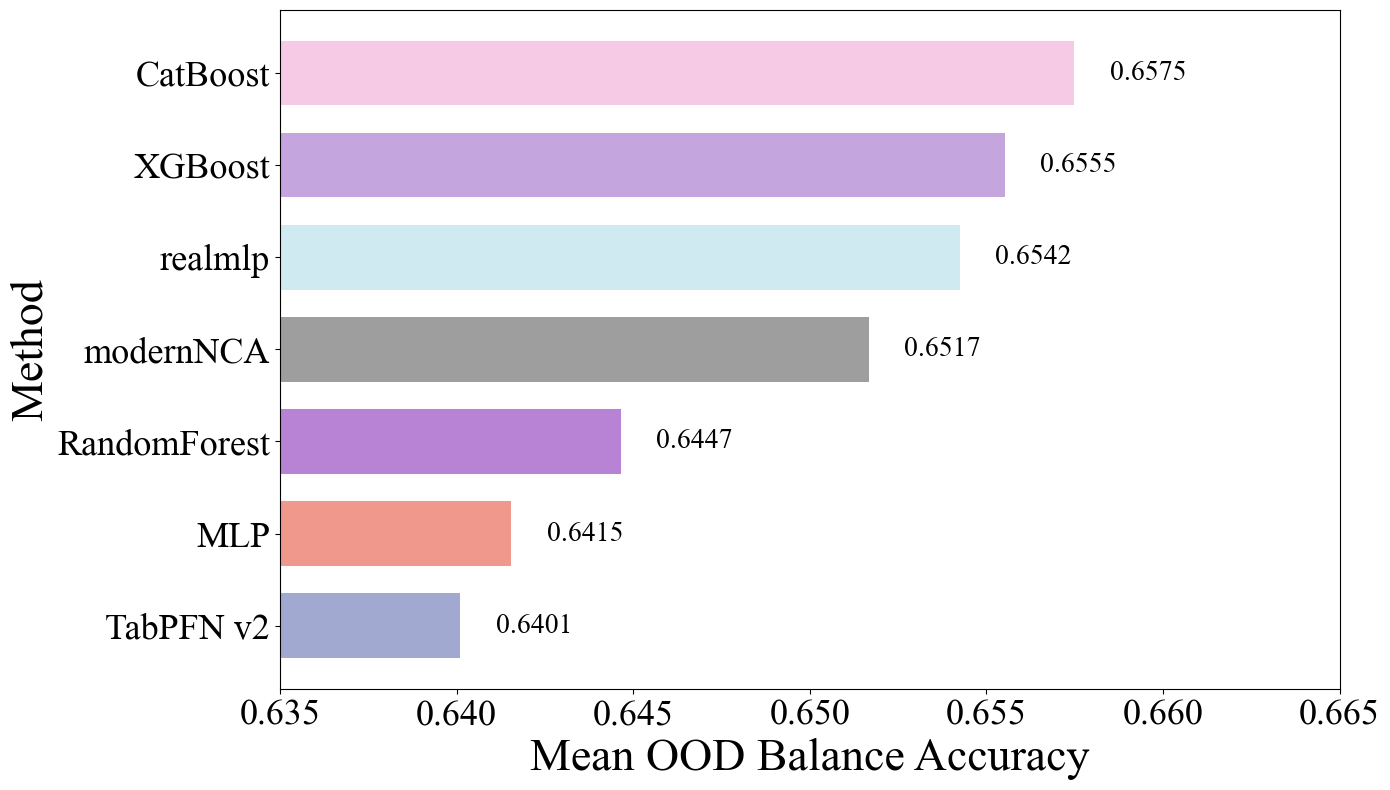}
  \caption{Mean Balance Accuracy Across 9 Datasets with Distribution Shifts}
  \label{fig:balance_acc}
\end{figure}

We evaluate TabPFN v2 alongside several mainstream tabular models under changing data distributions scenarios, using metrics including Accuracy, Balanced Accuracy, F1-score, and ROC-AUC. The evaluation is conducted on nine fully numerical datasets drawn from the WhyShift~\cite{liu2024need} and TableShift~\cite{gardner2024benchmarking} benchmarks, which contain three types of data distribution scenarios. To accommodate memory constraints and the current limitations of TabPFN in handling very large datasets, we apply stratified subsampling (up to 50,000 instances) while preserving the original train/test splits. Figure~\ref{image/acc.png}, \ref{fig:auc},~\ref{fig:f1}and~\ref{fig:balance_acc} demonstrates results on changing data distributions. Although there has been recent work attempting to apply TabPFN to scenarios involving temporal distribution shift~\cite{helli2024driftresilienttabpfnincontextlearning}, their implementation is not publicly available. Hence, we do not include this method in our evaluation.

\section{Varied Learning Objectives}
We conduct an analysis across four primary classification learning objectives: accuracy, ROC-AUC, F1-score, and Balanced Accuracy. The analysis is performed on i.i.d. datasets that are employed in Changing Data Distribution. Table~\ref{tab:objectives} demonstrates results on varied learning objectives.

\begin{table}[t]
\centering
\caption{Accuracy, ROC-AUC, F1-Score and Balanced-Accuracy on evaluated models. The best is in \textbf{bold} and the second best is \underline{underlined}.}
\label{tab:objectives}
\begin{center}
\begin{small}
\resizebox{\textwidth}{23mm}{
\begin{tabular}{ccccccccccc}
\toprule
Objective                          & \multicolumn{1}{c}{Model} & \multicolumn{1}{c}{ACS Income(CA-PR)}        & \multicolumn{1}{c}{ACS Mobility(MS-HI)}      & \multicolumn{1}{c}{ACS Pub.Cov(NE-LA)}       & \multicolumn{1}{c}{ACS Pub.Cov(2010-2017)}   & \multicolumn{1}{c}{ACS Income(Setting 21)}   & \multicolumn{1}{c}{ACS Income(Setting 22)}   & \multicolumn{1}{c}{college\_scorecard}       & \multicolumn{1}{c}{brfss\_diabetes}          & \multicolumn{1}{c}{diabetes\_readmission}    \\
\midrule
\multirow{7}{*}{Accuracy}         & XGBoost                   & \multicolumn{1}{c}{\underline{0.813}}    & \multicolumn{1}{c}{0.788}          & \multicolumn{1}{c}{0.819}          & \multicolumn{1}{c}{0.831}          & \multicolumn{1}{c}{0.845}          & \multicolumn{1}{c}{0.864}          & \multicolumn{1}{c}{\underline{0.949}}    & \multicolumn{1}{c}{0.872}          & \multicolumn{1}{c}{0.642}          \\
                                   & CatBoost                  & \multicolumn{1}{c}{\textbf{0.815}} & \multicolumn{1}{c}{\underline{0.801}}    & \multicolumn{1}{c}{\textbf{0.827}} & \multicolumn{1}{c}{\textbf{0.836}} & \multicolumn{1}{c}{\textbf{0.852}} & \multicolumn{1}{c}{\textbf{0.870}} & \multicolumn{1}{c}{\textbf{0.949}} & \multicolumn{1}{c}{0.874}          & \multicolumn{1}{c}{\underline{0.651}}    \\
                                   & MLP                       & \multicolumn{1}{c}{0.781}          & \multicolumn{1}{c}{0.760}          & \multicolumn{1}{c}{0.793}          & \multicolumn{1}{c}{0.794}          & \multicolumn{1}{c}{0.821}          & \multicolumn{1}{c}{0.843}          & \multicolumn{1}{c}{0.934}          & \multicolumn{1}{c}{0.869}          & \multicolumn{1}{c}{0.569}          \\
                                   & ModernNCA                 & \multicolumn{1}{c}{0.810}          & \multicolumn{1}{c}{0.800}          & \multicolumn{1}{c}{0.819}          & \multicolumn{1}{c}{0.822}          & \multicolumn{1}{c}{0.842}          & \multicolumn{1}{c}{0.859}          & \multicolumn{1}{c}{0.946}          & \multicolumn{1}{c}{0.874}          & \multicolumn{1}{c}{0.651}          \\
                                   & RandomForest              & \multicolumn{1}{c}{0.805}          & \multicolumn{1}{c}{0.800}          & \multicolumn{1}{c}{0.821}          & \multicolumn{1}{c}{0.823}          & \multicolumn{1}{c}{0.843}          & \multicolumn{1}{c}{0.859}          & \multicolumn{1}{c}{0.943}          & \multicolumn{1}{c}{\textbf{0.876}} & \multicolumn{1}{c}{0.649}          \\
                                   & RealMLP                   & \multicolumn{1}{c}{0.812}          & \multicolumn{1}{c}{0.796}          & \multicolumn{1}{c}{0.814}          & \multicolumn{1}{c}{0.819}          & \multicolumn{1}{c}{0.840}          & \multicolumn{1}{c}{0.856}          & \multicolumn{1}{c}{0.946}          & \multicolumn{1}{c}{\underline{0.875}}    & \multicolumn{1}{c}{\textbf{0.653}} \\
                                   \rowcolor{gray!30}& TabPFN v2                   & \multicolumn{1}{c}{0.806}          & \multicolumn{1}{c}{\textbf{0.804}} & \multicolumn{1}{c}{\underline{0.824}}    & \multicolumn{1}{c}{\underline{0.834}}    & \multicolumn{1}{c}{\underline{0.848}}    & \multicolumn{1}{c}{\underline{0.867}}    & \multicolumn{1}{c}{0.938}          & \multicolumn{1}{c}{0.875}          & \multicolumn{1}{c}{0.651}          \\
                                   \midrule
\multirow{7}{*}{ROC-AUC}           & XGBoost                   & 0.893                              & 0.815                              & 0.817                              & 0.790                              & 0.848                              & 0.865                              & \underline{0.978}                        & 0.807                              & 0.679                              \\
                                   & CatBoost                  & \textbf{0.897}                     & \underline{0.832}                        & \textbf{0.833}                     & 0.802                              & \textbf{0.861}                     & \textbf{0.877}                     & \textbf{0.978}                     & \underline{0.817}                        & \textbf{0.697}                     \\
                                   & MLP                       & 0.848                              & 0.758                              & 0.757                              & 0.721                              & 0.791                              & 0.807                              & 0.959                              & 0.809                              & 0.585                              \\
                                   & ModernNCA                 & 0.893                              & 0.825                              & 0.821                              & \underline{0.812}                        & 0.839                              & 0.855                              & 0.976                              & 0.812                              & 0.689                              \\
                                   & RandomForest              & 0.886                              & \textbf{0.833}                     & \underline{0.833}                        & \textbf{0.823}                     & 0.847                              & 0.861                              & 0.971                              & \textbf{0.817}                     & \underline{0.689}                        \\
                                   & RealMLP                   & \underline{0.893}                        & 0.806                              & 0.804                              & 0.801                              & 0.828                              & 0.845                              & 0.950                              & 0.808                              & 0.685                              \\
                                   \rowcolor{gray!30}& TabPFN v2                   & 0.888                              & 0.832                              & 0.831                              & 0.779                              & \underline{0.858}                        & \underline{0.873}                        & 0.968                              & 0.816                              & 0.688                              \\
                                   \midrule
\multirow{7}{*}{F1-Score}          & XGBoost                   & \underline{0.770}                        & 0.815                              & \underline{0.732}                        & 0.471                              & \underline{0.717}                        & \underline{0.693}                        & \textbf{0.792}                     & \textbf{0.239}                     & \underline{0.522}                        \\
                                   & CatBoost                  & \textbf{0.772}                     & \textbf{0.824}                     & \textbf{0.735}                     & 0.468                              & \textbf{0.719}                     & \textbf{0.694}                     & \underline{0.789}                        & \underline{0.210}                        & 0.517                              \\
                                   & MLP                       & 0.732                              & 0.790                              & 0.699                              & 0.434                              & 0.676                              & 0.645                              & 0.739                              & 0.157                              & 0.487                              \\
                                   & ModernNCA                 & 0.762                              & 0.817                              & 0.722                              & \textbf{0.646}                     & 0.651                              & 0.639                              & 0.778                              & 0.129                              & 0.507                              \\
                                   & RandomForest              & 0.749                              & 0.814                              & 0.709                              & 0.628                              & 0.626                              & 0.609                              & 0.748                              & 0.058                              & 0.446                              \\
                                   & RealMLP                   & 0.764                              & 0.813                              & 0.719                              & \underline{0.646}                        & 0.650                              & 0.636                              & 0.786                              & 0.182                              & \textbf{0.534}                     \\
                                   \rowcolor{gray!30}& TabPFN v2                   & 0.755                              & \underline{0.817}                        & 0.721                              & 0.416                              & 0.702                              & 0.675                              & 0.728                              & 0.044                              & 0.516                              \\
                                   \midrule
\multirow{7}{*}{Balanced-Accuracy} & XGBoost                   & 0.801                              & \underline{0.723}                        & \textbf{0.721}                     & 0.658                              & \textbf{0.739}                     & \textbf{0.741}                     & \underline{0.860}                        & \textbf{0.567}                     & 0.617                              \\
                                   & CatBoost                  & \textbf{0.807}                     & 0.720                              & \underline{0.715}                        & 0.656                              & \underline{0.733}                        & \underline{0.735}                        & 0.855                              & \underline{0.557}                        & \underline{0.622}                        \\
                                   & MLP                       & 0.775                              & 0.708                              & 0.703                              & 0.642                              & 0.718                              & 0.720                              & 0.839                              & 0.551                              & 0.558                              \\
                                   & ModernNCA                 & 0.801                              & 0.708                              & 0.702                              & \underline{0.685}                        & 0.705                              & 0.710                              & 0.849                              & 0.531                              & 0.619                              \\
                                   & RandomForest              & 0.791                              & 0.695                              & 0.687                              & 0.670                              & 0.685                              & 0.686                              & 0.814                              & 0.514                              & 0.604                              \\
                                   & RealMLP                   & \underline{0.803}                        & \textbf{0.723}                     & 0.714                              & \textbf{0.695}                     & 0.714                              & 0.716                              & \textbf{0.867}                     & 0.548                              & \textbf{0.628}      
                       \\
                                   \rowcolor{gray!30}& TabPFN v2                   & 0.794                              & 0.711                              & 0.704                              & 0.631                              & 0.718                              & 0.718                              & 0.810                              & 0.510                              & 0.621                              \\
                                   
\bottomrule
\end{tabular}}
\end{small}
\end{center}
\end{table}

\end{document}